\documentclass{article} 
\usepackage[final]{colm2026_conference}

\usepackage{microtype}
\usepackage{hyperref}
\usepackage{url}
\usepackage{booktabs}

\usepackage{float}
\usepackage{placeins}

\usepackage{latexsym}

\usepackage[T1]{fontenc}

\usepackage[utf8]{inputenc}

\usepackage{inconsolata}

\usepackage{graphicx}

\usepackage{amsmath}
\usepackage{amsfonts}
\usepackage{amssymb}
\usepackage{bbm}

\usepackage{url}            
\usepackage{booktabs}       
\usepackage{amsfonts}       
\usepackage{nicefrac}       
\usepackage{xcolor}         
\usepackage{pifont}

\usepackage{algorithm}
\usepackage[noend]{algpseudocode}
\usepackage{wrapfig}
\usepackage{caption}

\newcommand{\method}{\textsc{NSI}} 
\newcommand{\metric}{\textsc{NSI}}
\newcommand{\blimp}{\textsc{BLiMP}}

\usepackage{hyperref}       

\makeatletter
\@ifundefined{todo}{}{}
\makeatother

\usepackage{lineno}

\definecolor{darkblue}{rgb}{0, 0, 0.5}
\hypersetup{colorlinks=true, citecolor=darkblue, linkcolor=darkblue, urlcolor=darkblue}

\title{Grammatical ``grandmother neurons'' are rare in LLMs}

\author{Linyang He \qquad Nima Mesgarani \\
Zuckerman Mind Brain Behavior Institute, Columbia University\\
\texttt{linyang.he@columbia.edu} \\
}

\begin{document}

\ifcolmsubmission
   \linenumbers
\fi

\maketitle

\begin{abstract}
   Understanding how Large Language Models (LLMs) encode linguistic structures remains a fundamental challenge in interpretability research. While diagnostic classifiers (or "probes") are widely used for this task, they face significant methodological criticism: training auxiliary classifiers introduces capacity confounds and calibration issues, often making it difficult to distinguish the model's intrinsic representations from the probe's ability to learn the task. To address these limitations, we introduce a probe-free framework for localizing linguistic selectivity at the individual neuron level. Leveraging the controlled contrasts of linguistic minimal pairs, we propose a Neuron Separability Index (NSI), a metric that directly quantifies how reliably single neurons differentiate grammatical from ungrammatical constructions without parameter updates. Applying NSI across 68 linguistic paradigms and seven checkpoints reveals three main patterns: 1) raw separability reaches near-peak levels earlier for morphological and syntactic distinctions than for syntax–semantics interface and conceptual distinctions. 2) after permutation normalization, single-unit selectivity is sparse, weak, and narrowly tuned: only a small fraction of units are sensitive to an average paradigm, and strongly selective “grandmother neurons” are rare. 3) whole-vector linear separability, single-neuron selectivity, and behavioral competence are largely dissociated, and targeted ablations further separate activation selectivity from causal reliance.

\end{abstract}

\section{Introduction}
Understanding how large language models (LLMs) encode linguistic structure remains a central challenge for interpretability. A recurring question concerns \emph{granularity}: is a grammatical distinction carried by a small number of highly selective units, in the spirit of the long-debated ``grandmother neuron'' hypothesis in neuroscience \citep{kanwisher1997fusiform,gross2002genealogy,quiroga2005invariant,posani2025rarely}, or is it spread thinly across many units, each contributing only weakly? Work on \emph{superposition} and \emph{polysemanticity} suggests that neither extreme is guaranteed: single units can mix multiple features, so apparent selectivity may depend on which contrasts one happens to test \citep{elhage2022toy,bills2023language,huang2024ravel}. Sparse feature discovery can partially disentangle such mixtures, but the recovered directions are not automatically complete, atomic, or tied to model behavior, which makes unit-level functional claims difficult to establish \citep{bricken2023monosemanticity,cunningham2024sparse,makelov2025towards,leask2025sparse}. Settling the granularity question therefore requires a measurement that is defined at the level of individual units, comparable across many linguistic phenomena, and free of auxiliary training.

Existing evidence comes from three lines of work, none of which supplies such a measurement. Targeted minimal-pair evaluation established that language models capture a wide range of grammatical dependencies \citep{linzen2016assessing,marvin-linzen-2018-targeted,warstadt2020blimp,misra2023comps,jumelet2025multiblimp}, but it treats the model as a black box and says nothing about internal organization. Representation analyses open the box and map where linguistic information is decodable across depth \citep{tenney2019bert,starace2023probing,he2024decoding}, yet a trained probe reads out a \emph{whole activation vector} and contributes its own capacity, so its success is silent about whether any individual unit carries the distinction \citep{hewitt2019designing,pimentel2020information,belinkov2022probing}. Mechanistic interpretability does reach individual units \citep{lakretz2019emergence,geva2021transformer,finlayson2021causal,meng2022locating}, but each study is built around one or two phenomena, leaving open whether strong single-unit selectivity is the normal case or a rare exception across the grammar as a whole. We discuss all three lines in detail in Section~\ref{sec:related}.

The open question, then, is quantitative and comparative: \emph{if a model behaves grammatically, how much of that competence is expressed in individual neurons, and is such selectivity broad (domain-general) or narrow (domain-specific)?} We address it with a \emph{probe-free} neuron-level framework that exploits the controlled contrasts of minimal pairs. Combining BLiMP \citep{warstadt2020blimp} and COMPS \citep{misra2023comps}, we organize the suite into a three-level hierarchy of 4 domains, 13 phenomena, and 68 paradigms, so that selectivity can be compared on a common scale from a single agreement contrast up to entire linguistic domains. For each paradigm we feed matched grammatical and ungrammatical sentences into a language model, extract last-token activations, and, for each neuron independently, compute a correlation-derived \emph{raw separability} score between its paired positive and negative activation vectors. Because minimal pairs differ in as little as one token, this contrast is tightly controlled, and because the score is read directly off the activations, it requires no parameter updates and no auxiliary classifier.

Raw separability can still be inflated by lexical overlap or sampling noise. We therefore normalize it against a null distribution obtained by randomly swapping the grammatical labels within pairs, which preserves lexical content while destroying the grammatical contrast. The resulting \emph{Neuron Separability Index} (\metric{}) expresses each neuron's discrimination in units of its own null, making values comparable across neurons, layers, and paradigms: a high \metric{} indicates discrimination beyond what lexical content or noise explains, whereas a near-zero \metric{} indicates insensitivity to the targeted contrast.

\begin{figure*}[t]
   \centering
   \includegraphics[width=0.95\linewidth]{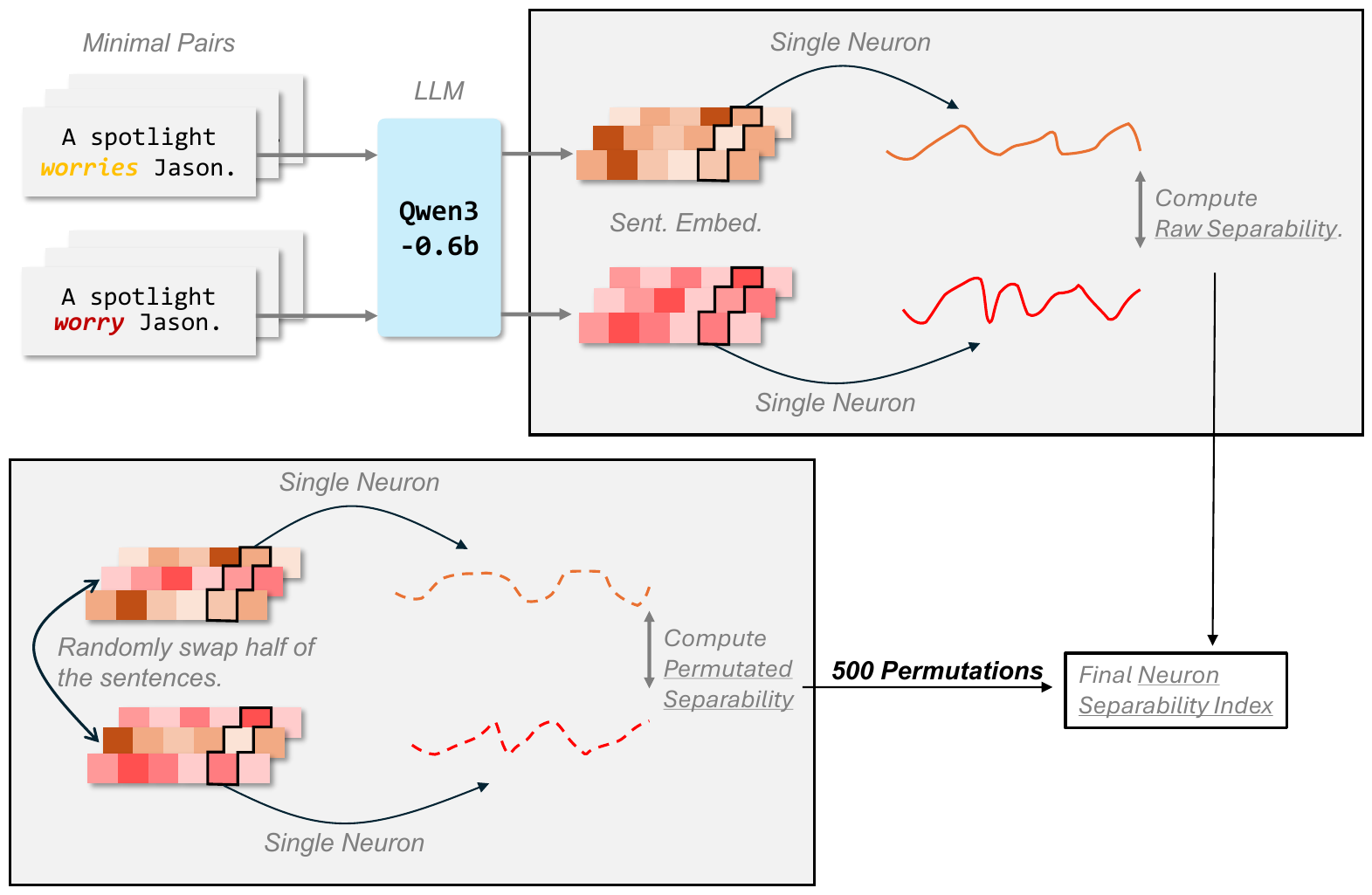}

   \caption{Minimal-pair neuron separability pipeline. Minimal pairs from BLiMP/COMPS are fed into a transformer LM (here, Qwen3-0.6B). For each minimal pair, we extract last-token activations, then focus on a single neuron and assemble paired activation vectors across items for the grammatical (positive) and ungrammatical (negative) sentences. We compute a raw separability score from the correlation between the two paired vectors. To control for lexical effects, we construct a null distribution by randomly swapping the grammatical/ungrammatical labels within half of the pairs and recomputing separability across many permutations; the observed raw score is then converted into a neuron separability index (NSI) value.}
   \label{fig:pipeline}
\end{figure*}

Applying this framework reveals a consistent but more nuanced picture. Raw separability reaches near-peak levels earlier for morphological and syntactic distinctions than for syntax--semantics interface and conceptual ones. After permutation normalization, however, the same activations yield a far more restrictive picture of single-unit selectivity: grammaticality-sensitive neurons are sparse, their effects are generally weak, and units meeting a strong-selectivity criterion are rare. Selectivity is also narrow, with even poly-selective neurons retaining a dominant within-domain preference rather than acting as general grammar detectors. This pattern replicates across seven checkpoints from four model families and survives our threshold and pairing controls.

Neuron-level selectivity also proves distinct from both whole-vector decodability and model behavior: probe accuracy is essentially unrelated to minimal-pair behavioral accuracy across paradigms, \metric{} is likewise only weakly associated with behavior, and targeted ablations of the rare above-threshold units and their highest-scoring same-layer neighbors are no more damaging than matched random ablations. Information can therefore be linearly decodable from a representation without being strongly localized to individual units, and neither property alone establishes that the model behaviorally relies on it. Our contributions are correspondingly threefold: a probe-free, permutation-normalized measure of unit-level selectivity; a systematic map of that selectivity across a three-level linguistic hierarchy; and an empirical dissociation between representational decodability, single-unit localization, and behavioral reliance.

\section{Minimal pair-based Neuron Separability}

We use minimal pairs from \blimp{} \citep{warstadt2020blimp} and COMPS \citep{misra2023comps} as a controlled testbed for grammatical and conceptual contrasts; full dataset details are provided in Appendix~\ref{sec:dataset}. Across the combined suite, the targeted contrasts are organized into a three-level hierarchy (Domain $\rightarrow$ Phenomenon $\rightarrow$ Paradigm), spanning four domains, 13 phenomena, and 68 paradigms in total. A domain is the broadest grouping, a phenomenon groups related minimal-pair paradigms, and a paradigm is one specific benchmark contrast.

Let $\mathcal{X}_p=\{(x_i^{+},x_i^{-})\}_{i=1}^{M_p}$ denote the minimal pairs for a BLiMP/COMPS paradigm $p$ (e.g., one specific subject--verb agreement contrast). For a transformer with $L$ layers and hidden width $d_\ell$ at layer $\ell$, let $h_{\ell}(x)\in\mathbb{R}^{d_\ell}$ be the last-token hidden state (for causal LMs). For neuron $j\in\{1,\dots,d_\ell\}$ define scalar activation $a_{\ell j}(x)=h_{\ell}(x)_j$.


\subsection{Raw separability score.}
For neuron $(\ell,j)$ in paradigm $p$, form paired activation vectors
\[
   \begin{aligned}
      \mathbf{v}^{+}_{\ell j,p} & = [a_{\ell j}(x^{+}_1),\ldots,a_{\ell j}(x^{+}_{M_p})], \\
      \mathbf{v}^{-}_{\ell j,p} & = [a_{\ell j}(x^{-}_1),\ldots,a_{\ell j}(x^{-}_{M_p})].
   \end{aligned}
\]
We first define the raw separability score $S^{\text{raw}}_{\ell j}(p)$ based on the correlation between the paired activation vectors:
\begin{equation}
   \label{eq:mpns_raw}
   S^{\text{raw}}_{\ell j}(p) = \frac{1-\mathrm{corr}\!\left(\mathbf{v}^{+}_{\ell j,p},\,\mathbf{v}^{-}_{\ell j,p}\right)}{2}.
\end{equation}
Intuitively, $S^{\text{raw}}_{\ell j}(p)$ is an \emph{effect-size} metric: it measures the degree of \emph{within-item contrast consistency} between grammatical and ungrammatical sentences in paradigm $p$, independent of any explicit null model. Pearson correlation performs its own centering and scale normalization, so no separately normalized activation variable is required. If a neuron responds differently to grammatical versus ungrammatical members across items, the correlation is low, yielding $S^{\text{raw}}_{\ell j}(p)$ close to~1.

To interpret Eq.~\eqref{eq:mpns_raw}, note that $S^{\text{raw}}_{\ell j}(p)\approx 0$ corresponds to nearly identical activation patterns across the paired sentences (corr$\approx 1$), $S^{\text{raw}}_{\ell j}(p)\approx 0.5$ corresponds to weak or inconsistent separation (corr$\approx 0$), and $S^{\text{raw}}_{\ell j}(p)\to 1$ corresponds to a strong, consistent contrast where the two conditions vary in opposite directions (corr$\to -1$).

\subsection{Permuted neuron separability index.}
To control for lexical effects and background noise, we employ a permutation test.
For each paradigm, we generate a null distribution by randomly swapping the grammatical/ungrammatical labels \emph{within} exactly half of the minimal pairs ($M_p/2$), while keeping the paired sentences themselves intact.
This procedure disrupts the systematic grammatical contrast while preserving the lexical content of each pair.
We repeat this process $N_{\text{perm}}=500$ times to obtain a distribution of permutation scores $\{S^{\text{perm}}_{k}\}_{k=1}^{N_{\text{perm}}}$.

In addition to the raw effect size $S^{\text{raw}}_{\ell j}(p)$, we define a neuron separability index metric that measures how strongly the observed separability exceeds the null distribution:
\begin{equation}
   \label{eq:mpns_z}
   \mathrm{NSI}_{\ell j}(p) = \max\left(0, \frac{S^{\text{raw}}_{\ell j}(p) - \mu_{\text{perm}}}{\sigma_{\text{perm}} + \epsilon}\right),
\end{equation}
where $\mu_{\text{perm}}$ and $\sigma_{\text{perm}}$ are the mean and standard deviation of the permutation scores, and $\epsilon$ is a small constant for numerical stability. We refer to $\mathrm{NSI}_{\ell j}(p)$ as \emph{Neuron Separability Index}. In contrast to the raw effect size, $\mathrm{NSI}_{\ell j}(p)$ is a \emph{noise-controlled} metric: it quantifies how strongly the observed separability exceeds the paradigm-matched null obtained by within-pair label swapping (i.e., beyond what can be explained by lexical content and sampling noise). In our later analyses, we retain \emph{both} separability metrics: $S^{\text{raw}}_{\ell j}(p)$ captures the direct separability effect size, while $\mathrm{NSI}_{\ell j}(p)$ provides a permutation-normalized, comparable measure of selectivity intensity across paradigms. Appendix~\ref{sec:nsi_semantics} characterizes what this score measures in terms of a neuron's item-wise response profile across matched items.

Because $\mathrm{NSI}_{\ell j}(p)$ is expressed in units of the permutation-null standard deviation, we use two operational thresholds in our analyses: neurons with $\mathrm{NSI}_{\ell j}(p){>}0$ are treated as \emph{sensitive} (showing positive separability), while neurons with $\mathrm{NSI}_{\ell j}(p){>}2$ are treated as \emph{strongly sensitive}. We use the latter as a descriptive strong-selectivity threshold rather than a Gaussian significance claim; Appendix~\ref{sec:threshold_calibration} reports a threshold sweep and empirical-null diagnostics. This criterion operationalizes a \textbf{``grandmother neuron''}: a single neuron whose activity alone can reliably discriminate grammatical vs.
ungrammatical minimal pairs for paradigm $p$.

\section{Experimental Setup}

Our primary analysis uses Qwen3-0.6B \citep{yang2025qwen3}. To test whether the main pattern is specific to this checkpoint, we repeat the analysis across Qwen3-1.7B, Qwen3-4B, Qwen3-8B, Pythia-410M, TinyLlama-1.1B, and Llama-3.1-8B \citep{dubey2024llama}. All models are decoder-only causal LMs, and we read the hidden state at the final non-padding token so that every representation has access to the complete sentence. Please refer to Appendix~\ref{sec:model_coverage} for architectures and coverage statistics.

We additionally evaluate robustness to the NSI threshold and the construction of sentence pairs, including random-pairing controls and critical-token deletion; please refer to Appendices~\ref{sec:threshold_calibration} and~\ref{sec:pairing_deletion} for the full protocols and results.

Finally, we compare neuron-level NSI with both model behavior and whole-vector linear probing on the 67 BLiMP paradigms, treating probing as a representation-level control rather than evidence of behavioral reliance. We also conduct a targeted ablation sanity check on the three paradigms containing an NSI${>}2$ unit: the candidates identified by the observational analysis are frozen, and at each candidate's own layer we zero the top $k\in\{5,10,20\}$ residual-stream dimensions at every non-padding position, comparing the resulting change in the minimal-pair log-probability margin, $\log p(x^+)-\log p(x^-)$, with bottom signed-score sets and 100 same-layer random sets. Please refer to Appendices~\ref{sec:probe_behavior} and~\ref{sec:targeted_ablation} for the complete setups, statistics, and results.

\subsection{Quantification of Hierarchical Tuning Breadth}
To characterize the functional specialization of neurons across network depths, we define ``tuning breadth'' ($B$) as the number of distinct categories to which a neuron responds at a specified level of the hierarchy. We analyze this metric at three levels of abstraction: fine-grained paradigms ($\mathcal{P}$), intermediate phenomena ($\mathcal{H}$), and broad domains ($\mathcal{D}$).

Let $z_{u,p}$ denote the selectivity score (NSI) of unit $u$ for paradigm $p$. We define the binary responsiveness indicator $r_{u,p} = \mathbbm{1}(z_{u,p} > 0)$, where $\mathbbm{1}(\cdot)$ is the indicator function. The tuning breadth for unit $u$ at each hierarchical level is calculated as follows:

\paragraph{Paradigm level.} The raw count of responsive paradigms is defined as:
\begin{equation}
   B^{(u)}_{\text{paradigm}} = \sum_{p \in \mathcal{P}} r_{u,p}
\end{equation}

\paragraph{Phenomenon and domain levels.} To account for the nested taxonomy, we apply a logical disjunction (Boolean OR) aggregation. A unit is considered responsive to a phenomenon $h$ or domain $d$ if it responds to \textit{at least one} constituent paradigm:
\begin{align}
   B^{(u)}_{\text{phenomenon}} & = \sum_{h \in \mathcal{H}} \left( \max_{p \in \mathcal{P}_h} r_{u,p} \right) \\
   B^{(u)}_{\text{domain}}     & = \sum_{d \in \mathcal{D}} \left( \max_{p \in \mathcal{P}_d} r_{u,p} \right)
\end{align}
where $\mathcal{P}_h$ and $\mathcal{P}_d$ denote the paradigms belonging to phenomenon $h$ and domain $d$, respectively.

Finally, the layer-wise expected tuning breadth $\bar{B}_k$ for layer $k$ is computed by averaging over all $N_k$ units in the layer:
\begin{equation}
   \bar{B}_k = \frac{1}{N_k} \sum_{u \in \text{Layer}_k} B^{(u)}
\end{equation}
This metric serves as a proxy for neuronal polysemanticity: higher values indicate broad generalization, while lower values indicate functional specialization.

\subsection{Clustering neurons by linguistic phenomenon selectivity}
\label{sec:cluster}
To analyze the functional organization of neurons, we move beyond individual paradigms and aggregate selectivity at the phenomenon level. Let $\mathcal{P}$ denote the set of tested paradigms and $\mathcal{H}$ the set of linguistic phenomena. Each phenomenon $h\in\mathcal{H}$ is associated with a subset $\mathcal{P}_h\subset\mathcal{P}$; for example, the Subject--Verb Agreement phenomenon contains several specific BLiMP paradigms. For a fixed layer $\ell$ (or after flattening neurons across the network), let $z_j(p)$ be the selectivity score of neuron $j$ for paradigm $p$. We define its phenomenon-level selectivity as the mean across constituent paradigms:
\[
v_{h,j}=\frac{1}{|\mathcal{P}_h|}\sum_{p\in\mathcal{P}_h}z_j(p).
\]

\paragraph{Neuron filtering via tuning breadth.}
Rather than selecting neurons based on peak magnitude alone, we retain neurons engaged across multiple phenomena. The phenomenon-level tuning breadth of neuron $j$ is
\[
B_j=\sum_{h\in\mathcal{H}}\mathbbm{1}(v_{h,j}>0).
\]
We select $\mathcal{N}^*=\{j\mid B_j\geq3\}$, focusing the clustering analysis on neurons with broader linguistic responsiveness rather than noise or paradigm-idiosyncratic tuning.

\paragraph{Clustering and matrix construction.}
We construct a phenomenon-by-neuron matrix $\mathbf{X}\in\mathbb{R}^{|\mathcal{H}|\times|\mathcal{N}^*|}$ from the aggregate scores of the selected neurons. Rows (phenomena) are grouped a priori by domain in the order \textit{Concept}, \textit{Syntax--Semantics Interface}, \textit{Syntax}, and \textit{Morphology}. We cluster the columns (neurons) to identify groups with similar tuning profiles. Specifically, we compute the pairwise correlation distance
\[
D(j,j')=1-\mathrm{corr}(\mathbf{X}_{:,j},\mathbf{X}_{:,j'}),
\]
and apply agglomerative hierarchical clustering with average linkage (UPGMA) \citep{schlee1975numerical}. The resulting dendrogram determines the column order in the heatmap, revealing groups of neurons with preferences for particular domains or phenomena.

\section{Results}
\label{sec:results}
\subsection{Raw Separability Patterns}
\addtolength{\columnsep}{-4pt}
\begin{wrapfigure}{r}[4pt]{0.5\textwidth}
   \vspace{-12pt}
   \centering
   \includegraphics[width=\linewidth]{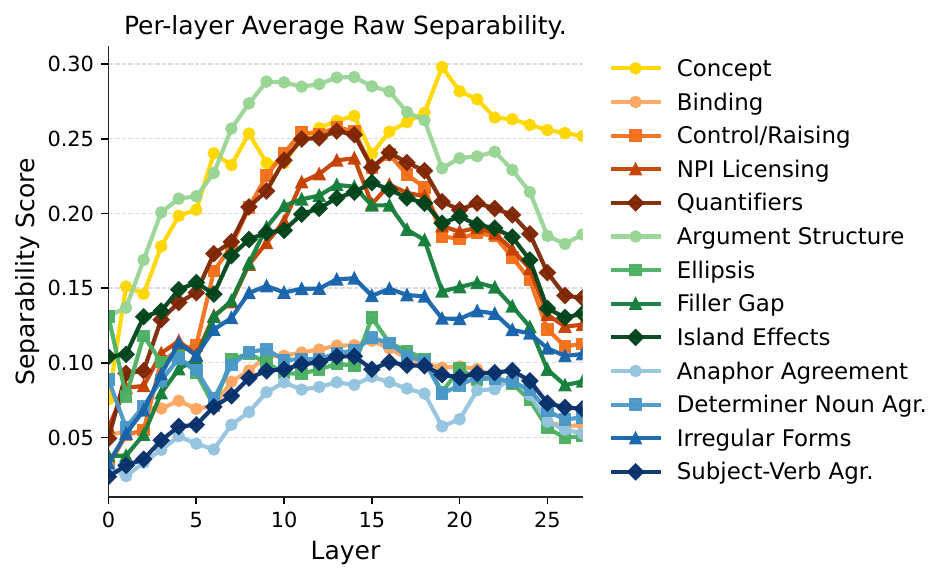}
   \captionsetup{skip=2pt}
   \caption{Colors indicate linguistic domains:
      \textcolor[HTML]{FFD700}{\textit{Concept}},
      \textcolor[HTML]{FF8C00}{\textit{Syntax--Semantics Interface}},
      \textcolor[HTML]{2E8B57}{\textit{Syntax}},
      and \textcolor[HTML]{4169E1}{\textit{Morphology}}. Layer-wise average separability curves across linguistic phenomena. Each curve reports mean neuron separability for one phenomenon, averaged across its constituent paradigms.
   }
   \label{fig:lms}
   \vspace{-12pt}
\end{wrapfigure}
\addtolength{\columnsep}{4pt}
Figure~\ref{fig:lms} shows that Syntax and Morphology concentrate around layer~14, whereas Concept peaks around layer~19, suggesting a shift from local grammatical constraints toward higher-level distinctions with depth. A complementary 90\% saturation analysis, computed separately for each of the 68 paradigms as the earliest layer reaching 90\% of its own peak and then averaged by domain, yields the same late-to-early ordering: Concept ($19.0$), Syntax--Semantics Interface ($10.3$), Syntax ($9.5$), and Morphology ($8.4$; Appendix Figure~\ref{fig:saturation_layers_by_phenomenon}). At the paradigm level, raw separability and whole-vector probing saturation layers show a positive correspondence (Spearman $\rho=0.23$; Appendix Figure~\ref{fig:raw_probe_paradigm_saturation_correspondence}), providing qualitative support for a shared coarse progression while showing substantial measure-specific variation.



\subsection{Neuron Separability Index Patterns}
\begin{figure*}[t]
   \centering
   \includegraphics[width=\linewidth]{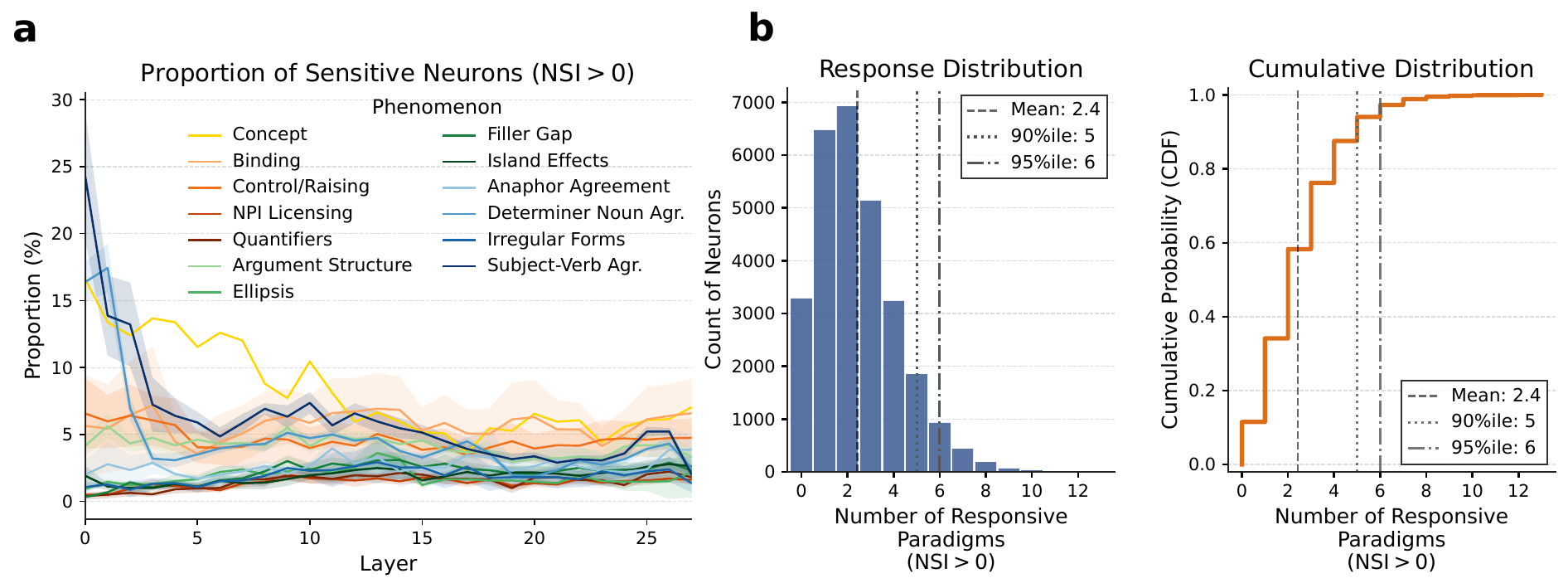}
   \caption{\textbf{(a)} Proportion of sensitive neurons ($\mathrm{NSI}{>}0$) across layers, broken down by linguistic phenomenon. Lines show the mean proportion across paradigms within each phenomenon; shaded bands indicate variability across paradigms. Overall, neurons with positive sensitivity to the targeted contrasts are sparse throughout the network. Mean recruitment differs across the four domains: Concept ($\approx 8.27\%$), Syntax--Semantics Interface ($\approx 3.31\%$), Syntax ($\approx 2.50\%$), and Morphology ($\approx 3.88\%$). Sensitivity also tends to decrease in deeper layers, suggesting that later layers allocate relatively less capacity to local grammatical well-formedness and more to higher-level information (e.g., semantics, reasoning, and discourse). The right panel of Appendix Figure~\ref{fig:proportion_positive} provides the paradigm-level distributions. \textbf{(b)} Distribution of \emph{neuron breadth} across paradigms. For each neuron (a unit at a specific layer), we count the paradigms for which it is responsive ($\mathrm{NSI}{>}0$). \textbf{Left:} histogram of the number of responsive paradigms per neuron. \textbf{Right:} empirical CDF of the same quantity. Vertical lines mark the mean and the 90th/95th percentiles, showing that most responsive neurons participate in only a small number of paradigms and that broad, multi-paradigm responsiveness is rare. Corresponding response distributions at the domain and phenomenon levels appear in Appendix Figures~\ref{fig:distribution_field} and~\ref{fig:distribution_form}.}
   \label{fig:sensitive_neurons}
\end{figure*}

\paragraph{Grammaticality sensitive neurons are sparse in LLMs.}
Figure~\ref{fig:sensitive_neurons}-(a) quantifies how many neurons exhibit positive sensitivity to grammatical contrasts (positive NSI) across layers.


Overall, such neurons constitute only a small fraction of the model at any layer, indicating that grammaticality is not broadly encoded by most neurons. This sparsity is also domain-dependent: the mean proportion of sensitive neurons is approximately $8.27\%$ for Concept, $3.31\%$ for Syntax--Semantics Interface, $2.50\%$ for Syntax, and $3.88\%$ for Morphology. Thus, Concept recruits the largest pool, whereas the three grammatical domains remain much sparser. In addition, the layer-wise trends reveal a clear decay: the proportion of grammar-sensitive neurons is highest in early to middle layers and gradually diminishes toward deeper layers. This pattern suggests a functional redistribution across depth. As representations become more abstract, fewer neurons are dedicated to encoding local grammatical well-formedness, and more capacity is likely devoted to higher-level information such as semantic integration, reasoning, and discourse context.

\begin{figure}[!t]
   \centering
   \includegraphics[width=\linewidth]{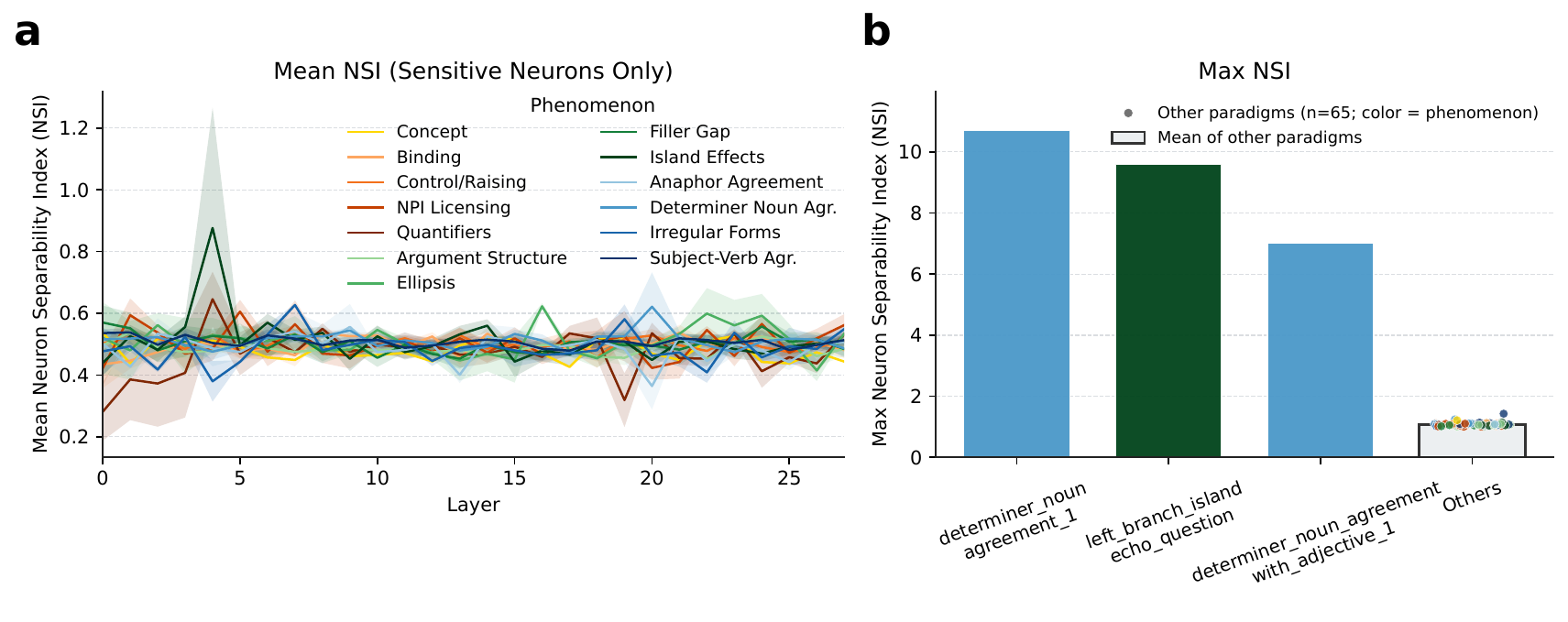}
   \captionsetup{skip=2pt}
   \caption{\textbf{(a)} Mean NSI across layers computed over sensitive neurons only (NSI${>}0$), grouped by linguistic phenomenon. Even within this subset, the average NSI remains small, indicating that sensitivity is generally weak rather than sharply selective. \textbf{(b)} For each BLiMP/COMPS paradigm, we compute the maximum NSI across neurons. Bar and point colors indicate each paradigm's parent phenomenon; the outlined \textit{Others} bar shows the mean over the remaining 65 paradigms. Only three paradigms contain a neuron with NSI${>}2$. Appendix Figure~\ref{fig:PNS_all} gives paradigm- and neuron-level detail.}
   \label{fig:no_grandma}
\end{figure}

\paragraph{``Grandmother neurons'' are rare in LLMs.}
One natural question is whether a given linguistic phenomenon is localized to a small set of ``grandmother'' neurons that respond strongly and selectively to that phenomenon. Using NSI, we find little evidence for such extreme localization. Figure~\ref{fig:no_grandma}a shows that even after restricting attention to sensitive neurons (NSI${>}0$), their average separability remains low (typically around $\sim0.5$). Consistently, for 65 of 68 paradigms, even the most selective neuron fails to reach NSI${>}2$, our operational threshold for strong selectivity. For each of the remaining three paradigms, Appendix Figure~\ref{fig:PNS_all} shows that the threshold is exceeded by only one neuron out of the full network ($28$ layers $\times$ $1024$ neurons).

Together, these results suggest that grammatical contrasts are typically encoded in a distributed manner: instead of a single neuron reliably separating grammatical from ungrammatical minimal pairs, separability is spread across neurons whose individual effects are modest.

\begin{wrapfigure}[21]{R}[4pt]{0.47\textwidth}
   \vspace{-8pt}
   \centering
   \includegraphics[width=\linewidth]{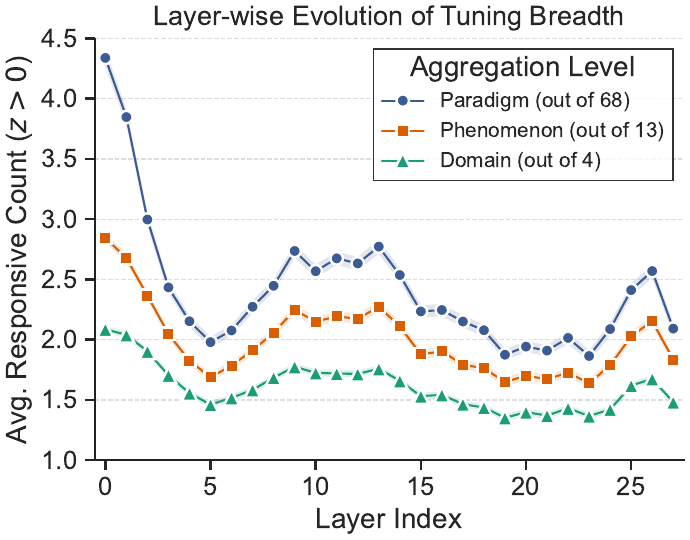}
   \captionsetup{skip=2pt}
   \caption{Layer-wise average tuning breadth at the paradigm, phenomenon, and domain levels. A neuron is counted as responsive to a phenomenon or domain when it has positive selectivity for at least one constituent paradigm. Shaded regions show standard error.}
   \label{fig:response_hierarchy}
   \vspace{-15pt}
\end{wrapfigure}

\paragraph{Most neurons are domain-specific and narrowly tuned.}
While Figure~\ref{fig:sensitive_neurons}a shows that only a small fraction of neurons are sensitive to any given paradigm, a complementary question is how many different paradigms a single neuron responds to. We define each neuron's breadth as the number of paradigms for which it has NSI${>}0$. Figure~\ref{fig:sensitive_neurons}b shows that most neurons respond to only a small number of paradigms and that broadly responsive neurons are rare.

Figure~\ref{fig:response_hierarchy} shows the average number of responsive paradigms, phenomena, and domains per neuron. Breadth declines sharply in the early layers, remains low through the middle of the network, and increases modestly near the output. Together with the scarcity of NSI${>}2$ units, this pattern supports a representation in which grammatical information is distributed across neurons with narrow, weak-to-moderate effects rather than concentrated in a few universally responsive units.

\FloatBarrier

\begin{figure*}[t]
   \centering
   \includegraphics[width=1.0\linewidth]{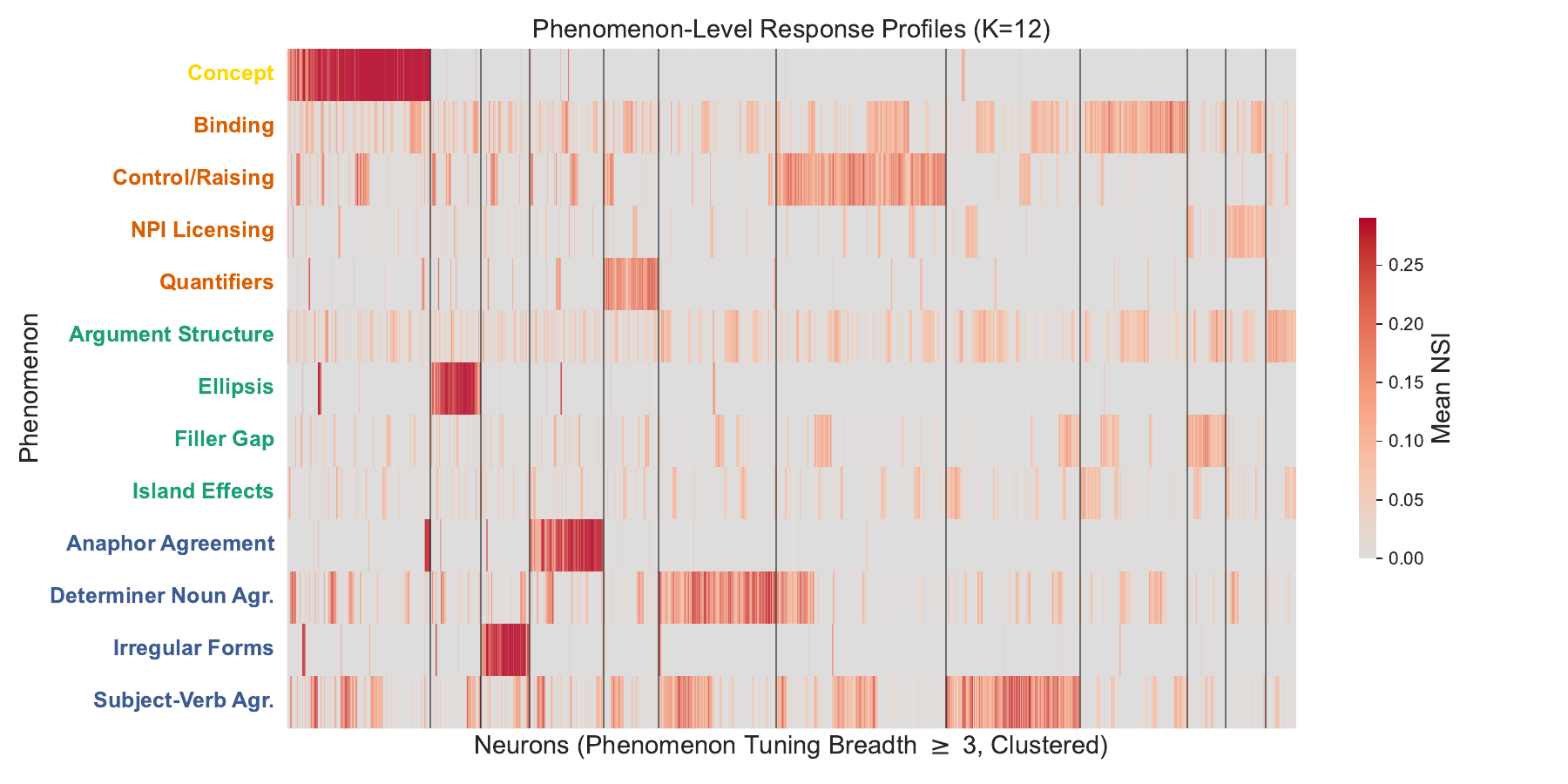}
   \caption{Functional modularity in tuning across linguistic phenomena. The heatmap displays mean phenomenon-level NSI for neurons responsive to at least three phenomena. Rows are grouped by domain, and columns are hierarchically clustered neurons. The block structure shows that even poly-selective neurons retain dominant within-domain preferences rather than behaving as uniform cross-domain grammar detectors.}
   \label{fig:heatmap_cluster}
\end{figure*}

\paragraph{Even broadly tuned neurons remain selective.}
Some neurons are poly-selective in the sense that they respond to at least three linguistic phenomena, but Figure~\ref{fig:heatmap_cluster} shows that this does not make them general-purpose grammar neurons. Clustering reveals clear blocks in which a unit has comparatively strong mean NSI for one phenomenon or a small group of related phenomena, while its responses elsewhere are weaker. Cross-phenomenon co-activation therefore resembles spillover around a dominant preference rather than uniformly strong selectivity across domains.

\subsection{Robustness and Generalization}

\begin{figure*}[!t]
   \centering
   \includegraphics[width=\linewidth]{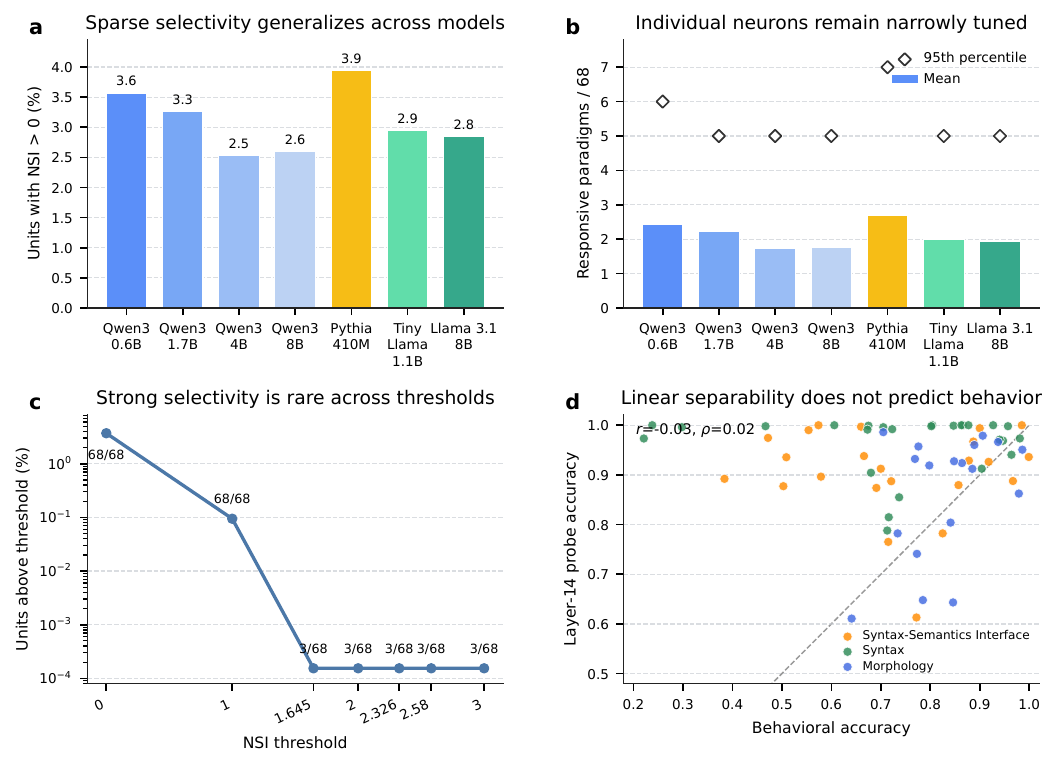}
   \captionsetup{skip=2pt}
   \caption{Robustness and generalization of neuron-level selectivity. \textbf{(a)} Across four Qwen3 scales and three additional checkpoints, only $2.5$--$3.9\%$ of units have positive NSI for an average paradigm. \textbf{(b)} The average neuron responds to only $1.7$--$2.7$ of 68 paradigms; diamonds mark the 95th percentile. \textbf{(c)} For Qwen3-0.6B, the fraction of units above threshold falls sharply, and only 3 of 68 paradigms contain any unit from NSI $\geq 1.645$ through NSI $>3$. \textbf{(d)} Whole-vector probe accuracy is essentially uncorrelated with minimal-pair behavior across 67 BLiMP paradigms.}
   \label{fig:robustness_generalization}
   \vspace{-8pt}
\end{figure*}

\paragraph{Generalization across models.}
The sparse, narrow-tuning pattern is not specific to Qwen3-0.6B. Across Qwen3-0.6B/1.7B/4B/8B, Pythia-410M, TinyLlama-1.1B, and Llama-3.1-8B, an average paradigm recruits only $2.52$--$3.94\%$ of units with positive NSI (Figure~\ref{fig:robustness_generalization}a). Viewed from the complementary per-neuron perspective, mean tuning breadth ranges from only $1.72$ to $2.68$ responsive paradigms out of 68, and the 95th percentile is $5$--$7$ paradigms (Figure~\ref{fig:robustness_generalization}b). Each checkpoint has only $2$--$6$ paradigms with any NSI${>}2$ unit. Across model scales and families, this sparse-selectivity pattern remains consistent. Complete model statistics and breadth distributions are in Appendix~\ref{sec:model_coverage}.

\paragraph{Robustness to analysis choices.}
The conclusion does not depend on the exact cutoff. At NSI${>}1$, all paradigms retain at least one above-threshold unit, yet these units comprise only $0.094\%$ of the network on average. From NSI${}\geq1.645$ through NSI${>}3$, only 3 of 68 paradigms contain any above-threshold unit, with one such unit per paradigm (Figure~\ref{fig:robustness_generalization}c). Pairing controls clarify why both item matching and null normalization matter: random good--good and bad--bad pairs have raw separability near $0.50$ but yield no NSI${>}2$ units, whereas breaking item matching in cross-item good--bad pairs creates sporadic high-NSI artifacts. In the clean one-prefix deletion subset, removing the critical contrast token reduces mean raw separability from $0.0753$ to $0.0011$, while deleting a matched non-critical token leaves it at $0.0746$. Appendix~\ref{sec:threshold_calibration} reports the complete threshold analysis, and Appendix~\ref{sec:pairing_deletion} reports the complete pairing controls.

\paragraph{Representation is not behavioral reliance.}
Whole-vector representations are highly linearly separable (mean layer-14 probe accuracy $0.917$), while the same model's mean BLiMP behavioral accuracy is $0.758$. Across paradigms the two are effectively unrelated (Pearson $r=-0.032$; Figure~\ref{fig:robustness_generalization}d). For example, \texttt{sentential\_subject\_island} has perfect probe accuracy ($1.000$) but behavioral accuracy of only $0.238$. Thus, whole-vector linear separability, single-neuron NSI, and behavioral preference are empirically distinct quantities; neither probing nor NSI alone establishes causal use. Appendix~\ref{sec:probe_behavior} provides the full correlation, domain-level, and paradigm-level results.

\paragraph{Targeted group ablation does not reveal NSI-specific effects.}
We use ablations of $k\in\{5,10,20\}$ dimensions as the primary intervention, with each top group containing the frozen NSI${>}2$ candidate and its highest-scoring same-layer neighbors (Table~\ref{tab:ablation_groups}). Across all nine group comparisons, no top set is more damaging than 100 same-size random sets at $p<0.05$ (all empirical $p\geq0.109$), and absolute accuracy changes remain at or below $0.009$. Both agreement paradigms show near-zero or positive margin changes across group sizes. The largest targeted drop occurs for the left-branch island paradigm at $k=20$ ($\Delta M=-0.543$, paired 95\% CI $[-0.630,-0.455]$), but it is not selective relative to random groups ($p=0.109$), and the corresponding bottom-20 control also reduces the margin ($-0.439$). The group intervention therefore does not establish that high-NSI dimensions form a behaviorally privileged subset. This pattern is consistent with distributed or redundant encoding, although the null selective result does not prove that the selected dimensions are irrelevant. Appendix~\ref{sec:targeted_ablation} reports the random intervals, top-1 auxiliary check, and all conditions.

\begin{table*}[!t]
   \centering
   \scriptsize
   \setlength{\tabcolsep}{5pt}
   \renewcommand{\arraystretch}{0.92}
   \captionsetup{skip=2pt}
   \caption{Primary targeted group-ablation results. $\Delta M$ is the change in mean $\log p(x^+)-\log p(x^-)$; brackets give paired-bootstrap 95\% CIs, and $p_{\mathrm{rand}}$ compares each top group with 100 same-size random groups.}
   \label{tab:ablation_groups}
   \begin{tabular}{@{}lrcrrr@{}}
      \toprule
      Paradigm & $k$ & Top $\Delta M$ [95\% CI] & Random mean & Bottom $\Delta M$ & $p_{\mathrm{rand}}$ \\
      \midrule
      Det.--noun agr. & 5  & $-0.005$ [$-0.014, 0.003$] & $-0.003$ &  0.147 & 0.446 \\
                       & 10 & $ 0.000$ [$-0.011, 0.011$] & $-0.009$ &  0.054 & 0.634 \\
                       & 20 & $ 0.111$ [$ 0.084, 0.138$] & $-0.034$ & $-0.022$ & 0.911 \\
      \midrule
      Det.--noun agr.+adj. & 5  & $-0.006$ [$-0.015, 0.004$] & $-0.005$ & 0.161 & 0.545 \\
                            & 10 & $-0.005$ [$-0.020, 0.012$] & $-0.012$ & 0.177 & 0.604 \\
                            & 20 & $ 0.026$ [$ 0.004, 0.048$] & $-0.009$ & 0.351 & 0.733 \\
      \midrule
      Left-branch island & 5  & $ 0.009$ [$-0.019, 0.038$] & $-0.077$ & $-0.229$ & 0.752 \\
                          & 10 & $ 0.001$ [$-0.039, 0.040$] & $-0.166$ & $-0.488$ & 0.693 \\
                          & 20 & $-0.543$ [$-0.630,-0.455$] & $-0.148$ & $-0.439$ & 0.109 \\
      \bottomrule
   \end{tabular}
   \vspace{-8pt}
\end{table*}

\section{Conclusion}
We present \method{}, a minimal-pair, probe-free neuron-level measure that localizes linguistic selectivity across layers and characterizes it consistently at the domain, phenomenon, and paradigm levels. The approach is simple, statistically principled, and complementary to mechanistic analyses. Targeted ablations of groups containing the three rare NSI${>}2$ candidates and their highest-scoring same-layer neighbors yield no effect distinguishable from same-size random controls, reinforcing the distinction between activation selectivity and causal reliance. Future work will move beyond residual-coordinate ablation to connect neuron clusters to concrete causal circuits through patching and mediation analysis \citep{vig2020investigating}, and extend to multilingual minimal pairs.

\FloatBarrier

\bibliography{references,references_codex}
\bibliographystyle{colm2026_conference}

\appendix
\section*{Limitations}
NSI is a probe-free lens on neuron-level selectivity, but the results should be interpreted within the following scope.

\paragraph{Models and data.}
We study seven decoder-only checkpoints from the Qwen3, Pythia, TinyLlama, and Llama families; differences in corpora, tokenizers, architectures, objectives, or post-training could change sparsity and apparent localization. BLiMP and COMPS are controlled, templated English resources and under-represent natural discourse, pragmatics, gradient acceptability, and typologically diverse languages. The conclusions are therefore robust within the tested models and benchmarks, not universal.

\paragraph{Separability is not causality.}
NSI quantifies activation separability, while our intervention tests whether selected residual coordinates have uniquely strong causal effects. These within-benchmark tests cover three paradigms in one checkpoint; held-out phenomena and circuit-level interventions such as activation patching and mediation would provide stronger generalization and causal tests.

\paragraph{Representation choices.}
We use last-token layer outputs and treat each scalar dimension as a neuron. This can miss information at earlier positions, distributed over time, or expressed in attention and MLP signals; neuron definitions also vary across architectures. Alternative positions, components, and readouts may yield different selectivity patterns.

\paragraph{Calibration and cost.}
We use a matched within-pair permutation null that preserves lexical content while disrupting grammatical labels. Future work can model remaining lexical, tokenization, template, or item dependencies more explicitly. Robust normalization requires extensive permutation testing, but the independent permutations can be parallelized. Different similarity metrics emphasize distinct properties of neuronal responses. Although the sparse unit-level pattern persists under the Spearman and cosine variants (Appendix~\ref{sec:threshold_calibration}), future work could more systematically examine what metric-dependent differences reveal about neuronal response structure and linguistic selectivity.

\clearpage
\section{Related Work}
\label{sec:related}

\paragraph{Targeted syntactic evaluation and minimal pairs.}
Evaluating the grammatical competence of language models has evolved from calculating overall perplexity to using targeted diagnostic datasets. Early work introduced small-scale, hand-crafted test suites to verify specific syntactic generalizations \citep{linzen2016,gulordava2018,marvin-linzen-2018-targeted,wilcox2018rnn}. This methodology was significantly scaled up with benchmarks like BLiMP \citep{warstadt2020blimp} and automated platforms such as SyntaxGym \citep{gauthier2020syntaxgym}, which use minimal pairs to isolate grammatical phenomena and reveal systematic gaps in LM performance, and was subsequently broadened to systematic syntactic generalization suites and to conceptual and multilingual coverage \citep{hu2020systematic,mueller-etal-2020-cross,liu2024zhoblimp,he2025xcomps}. High benchmark accuracy is nonetheless difficult to interpret on its own, as it can coexist with instability, prompt sensitivity, and divergence from human grammatical judgments \citep{dentella2023systematic,hu2023prompting,mahowald2024dissociating}. More fundamentally, while these behavioral metrics effectively diagnose \emph{what} linguistic rules a model violates, they treat the model as a black box, offering limited insight into \emph{where} and \emph{how} these distinctions are represented internally \citep{he2024decoding,he2025large}.

\paragraph{Representational structure and probing.}
To understand internal representations, the community turned to diagnostic classifiers, or ``probes.'' Seminal layer-wise analyses demonstrated that classical NLP pipeline steps, such as part-of-speech tagging and parsing, are naturally rediscovered in the hierarchical geometry of transformer representations \citep{tenney2019bert,hewitt2019structural,manning2020emergent,de2020s,koto2021discourse,he2025layer,he2025far,ju2024large}, although the reported ordering varies with architecture, task, and metric \citep{belinkov2019analysis,rogers2020primer}. Despite these insights, the probing paradigm faces significant methodological criticism. A core debate concerns whether a probe reveals the model's intrinsic knowledge or merely exploits the probe's own capacity to learn the task from the embeddings \citep{hewitt2019designing}. Theoretical work using information-theoretic criteria further suggests that probing results can be confounded by the ease of extracting information rather than its explicit presence \citep{pimentel2020information,voita2020information}, and related analyses show that probes can succeed on information the model does not actually use and that probe rankings are sensitive to design choices \citep{ravichander2021probing,kunz2022does}. These limitations motivate the need for \emph{probe-free} diagnostics, like our proposed framework, which directly measure selectivity without the interference of auxiliary training.

\paragraph{Mechanistic interpretability and neuron-level analyses.}
Mechanistic interpretability instead attributes computations to concrete components, such as attention heads, feed-forward key--value memories, and circuits supporting factual recall \citep{elhage2021mathematical,geva2021transformer,meng2022locating}. At the finest granularity, neuron-level studies have linked individual units to translation and morphology features \citep{bau2018identifying,dalvi2019one} and to factual knowledge \citep{dai-etal-2022-knowledge}, while methodological work cautions that probe-based neuron ranking conflates information that is encoded with information the model actually uses \citep{antverg2022on}. Closest to our setting, causal mediation has localized neuron-level contributions to subject--verb agreement in English \citep{finlayson2021causal} and across multilingual models \citep{mueller2022causal}, although the reliability of such interventions depends on patching design choices and their documented failure modes \citep{zhang2023towards}. These studies show that unit-level analysis is feasible, but each targets one or two phenomena in isolation, leaving open how typical their findings are. Our study is systematic in both coverage and measurement: a single classifier-free index is applied uniformly to 68 paradigms nested under 13 phenomena and four domains, computed identically for every neuron in every layer, and replicated across seven checkpoints from four model families under a common set of pairing and threshold controls. To our knowledge, this is the first survey of neuron-level linguistic selectivity at this scale, and it shifts the question from whether selective neurons exist for a given construction to how common and how broad such selectivity is across the grammar as a whole.


\clearpage
\section{Dataset}
\label{sec:dataset}
We leverage two complementary minimal-pair resources.

\paragraph{COMPS.} The \textbf{COMPS} dataset \citep{misra2023comps} extends minimal-pair evaluation to conceptual and semantic compositional phenomena, focusing on whether models (and individual neurons) track meaning-sensitive distinctions beyond surface form. COMPS pairs are constructed to preserve as much lexical overlap as possible while flipping a compositional or conceptual requirement. We pool the selected COMPS contrasts into one paradigm, \texttt{comps\_base}, assigned to the Concept phenomenon within the Concept domain. One example is:

(i) \textbf{Domain: Concept; phenomenon: Concept} \\
\hspace*{2em}\textit{a) \hspace*{0.5em}A \underline{kettle} is used for boiling.}  \\
\hspace*{2em}\textit{b) *A \underline{hammer} is used for boiling.}

\paragraph{BLiMP.} The \textbf{BLiMP} benchmark \citep{warstadt2020blimp} contains 67 paradigms of automatically generated minimal pairs, where each pair differs minimally but flips grammatical acceptability. We group these paradigms into 12 phenomena nested within three domains: \textbf{Syntax--Semantics Interface} (Binding, Control/Raising, NPI Licensing, and Quantifiers), \textbf{Syntax} (Argument Structure, Ellipsis, Filler Gap, and Island Effects), and \textbf{Morphology} (Anaphor Agreement, Determiner--Noun Agreement, Irregular Forms, and Subject--Verb Agreement).
Examples from paradigms in each BLiMP domain include:

(ii) \textbf{Domain: Syntax--Semantics Interface; phenomenon: NPI Licensing}\\
\hspace*{2em}\textit{a) \hspace*{0.5em}Even Suzanne has \underline{really} joked around.}  \\
\hspace*{2em}\textit{b) *Even Suzanne has \underline{ever} joked around.}

(iii) \textbf{Domain: Syntax; phenomenon: Filler Gap}\\
\hspace*{2em}\textit{a) \hspace*{0.5em}Mark figured out \underline{that} most governments appreciate Steve.} \\
\hspace*{2em}\textit{b) *Mark figured out \underline{who} most governments appreciate Steve.}

(iv) \textbf{Domain: Morphology; phenomenon: Subject--Verb Agreement}\\
\hspace*{2em}\textit{a) \hspace*{0.5em}The \underline{hospital} appreciates Claire.} \\
\hspace*{2em}\textit{b) *The \underline{hospitals} appreciates Claire.}

Together, the combined hierarchy contains four domains, 13 phenomena, and 68 paradigms: 67 BLiMP paradigms plus one pooled COMPS paradigm. The two resources provide $116{,}300$ English minimal pairs generated from linguist-crafted templates, with 96.4\% (BLiMP) and 93.1\% (COMPS) human agreement.

\FloatBarrier
\section{Models}
\label{sec:models}
We conduct our main analysis on the recent \textbf{Qwen3-0.6B} model \citep{yang2025qwen3}, a 0.6-billion-parameter causal transformer released by Alibaba Group. Qwen3 adopts a decoder-only architecture with rotary position embeddings, multi-head self-attention, and feed-forward layers following the standard transformer design. Despite its relatively modest size, Qwen3-0.6B achieves strong performance across a wide range of language modeling and reasoning benchmarks, making it a suitable testbed for fine-grained interpretability studies. Its compact scale allows us to efficiently extract and analyze neuron-level activations across all layers.

To study the generality of neuron-level selectivity, we additionally analyze larger Qwen3 checkpoints (\textbf{Qwen3-1.7B}, \textbf{Qwen3-4B}, and \textbf{Qwen3-8B}) and three checkpoints from other model families: \textbf{Pythia-410M}, \textbf{TinyLlama-1.1B}, and \textbf{Llama-3.1-8B} \citep{dubey2024llama}. Architecture, calibration, sparsity, and tuning-breadth statistics are reported in Appendix~\ref{sec:model_coverage}.

\section{What does NSI measure?}
\label{sec:nsi_semantics}

Because $S^{\mathrm{raw}}$ is a function of the Pearson correlation between the paired activation vectors $\mathbf{v}^{+}$ and $\mathbf{v}^{-}$, it characterizes a neuron's \emph{item-wise response profile}, that is, the pattern of relative responses across the matched items of a paradigm. It is therefore informative about how the grammatical contrast reorganizes that profile, and is by construction invariant to the overall level and scale of the neuron's response.

The affine case makes this behavior explicit. Suppose the grammatical manipulation acts on a neuron as
\begin{equation}
   \label{eq:nsi_affine}
   \mathbf{v}^{-} = a\,\mathbf{v}^{+} + c\,\mathbf{1},
\end{equation}
with gain $a \neq 0$, constant offset $c$, and $\mathbf{1}$ the all-ones vector. Then $\operatorname{corr}(\mathbf{v}^{+},\mathbf{v}^{-}) = \operatorname{sign}(a)$, so
\[
   S^{\mathrm{raw}} =
   \begin{cases}
      0, & a > 0,\\
      1, & a < 0.
   \end{cases}
\]
A uniform offset ($a=1$, $c \neq 0$) and a pure gain change ($a > 0$, $c = 0$) both leave the relative ordering of items intact and receive the minimum score, whereas an order-reversing response ($a < 0$) receives the maximum. Departures from an order-preserving affine relation fall between these limits, with the score increasing as the item-wise profile under the ungrammatical condition departs further from its grammatical counterpart.

The permutation normalization then asks whether the observed reorganization exceeds what within-pair label swapping produces. A high NSI therefore indicates that the grammatical contrast systematically reshapes a neuron's relative response profile across matched linguistic items, beyond the permutation null.

This is the quantity the minimal-pair design is best suited to identify. Because the two members of a pair differ by as little as one token, some change in a neuron's overall activation level is expected for a large fraction of units and reflects the lexical substitution as much as the grammatical contrast itself. A change in the relative ordering across items is more specific, in that it indicates that the neuron's response depends on the grammatical status of each individual item. NSI is accordingly a measure of item-specific profile selectivity; mean-level effects constitute a complementary aspect of the paired contrast, which the uncentered cosine variant in Appendix~\ref{sec:threshold_calibration} retains.

\clearpage
\FloatBarrier
\section{Raw-Score Layer Localization}
\label{sec:raw_layer_localization}
\begin{figure*}[h]
   \centering
   \includegraphics[width=0.90\textwidth]{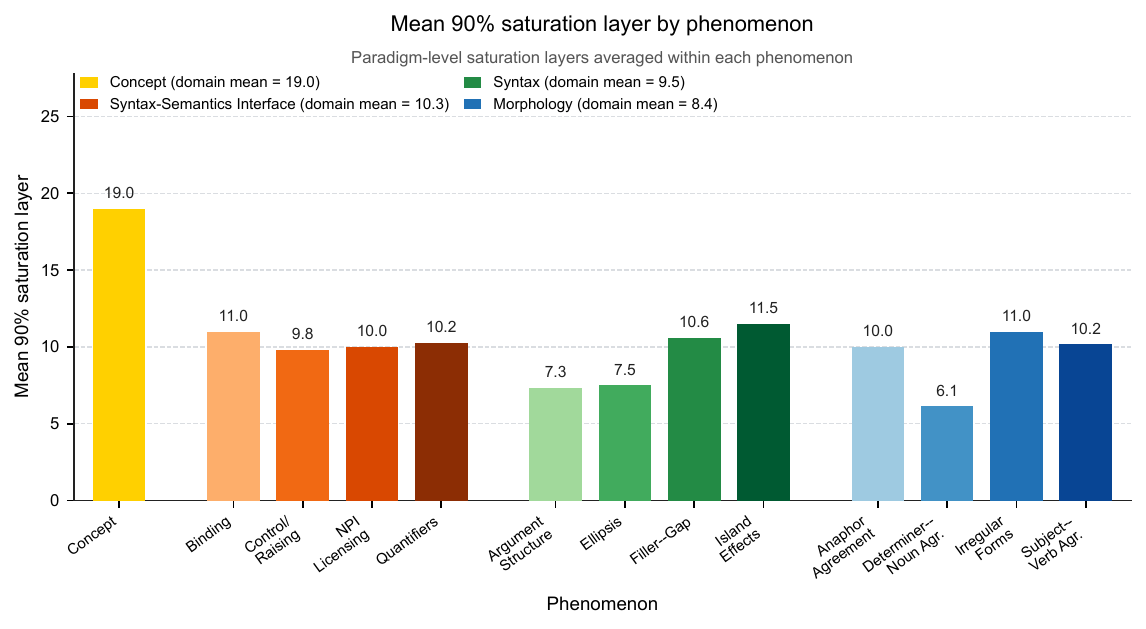}
   \caption{Paradigm-first 90\% saturation layers summarized by phenomenon. For each paradigm, we identify the earliest layer at which its median raw-separability score reaches 90\% of its own maximum. Bars show the mean of these paradigm-level saturation layers within each phenomenon. The legend reports means computed directly over all constituent paradigms in each domain: 19.0 for Concept, 10.3 for Syntax--Semantics Interface, 9.5 for Syntax, and 8.4 for Morphology.}
   \label{fig:saturation_layers_by_phenomenon}
\end{figure*}

\begin{figure*}[h]
   \centering
   \includegraphics[width=0.62\textwidth]{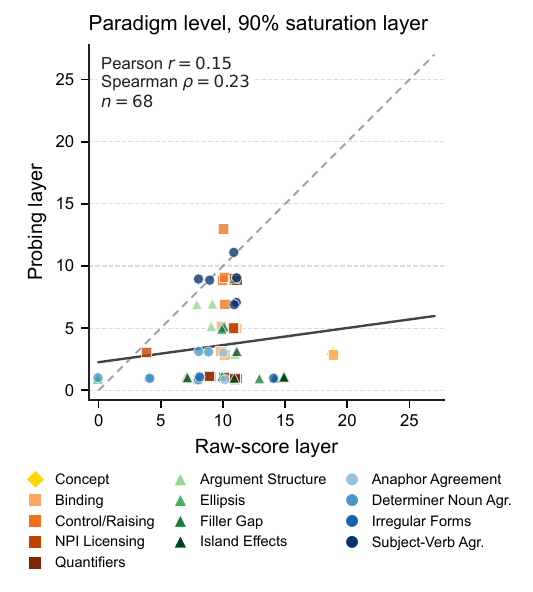}
   \caption{Correspondence between raw-separability and probing 90\% saturation layers across the 68 paradigms. For each paradigm and measure, saturation is the earliest layer reaching 90\% of that paradigm's own maximum. Colors encode phenomena, marker shapes encode domains, the dashed diagonal indicates equal layers, and the solid line is the least-squares fit. The association is weakly positive (Pearson $r=0.15$; Spearman $\rho=0.23$), indicating limited paradigm-level agreement between the two localization measures. Probing embedding index 0 is excluded, and sampled hidden-state indices 2, 4, \ldots, 28 are mapped to model layers 1, 3, \ldots, 27.}
   \label{fig:raw_probe_paradigm_saturation_correspondence}
\end{figure*}

We first report localization patterns from the raw separability score, following the same raw-score-first order as the main results. Figure~\ref{fig:saturation_layers_by_phenomenon} summarizes the paradigm-first saturation analysis, and Figure~\ref{fig:raw_probe_paradigm_saturation_correspondence} provides a cross-method comparison with whole-vector probing.

\clearpage
\FloatBarrier
\section{NSI Selectivity and Tuning Breadth}
\label{sec:nsi_selectivity_breadth}

\subsection{Cross-model robustness}
\label{sec:model_coverage}
\begin{figure}[H]
   \centering
   \includegraphics[width=0.88\linewidth]{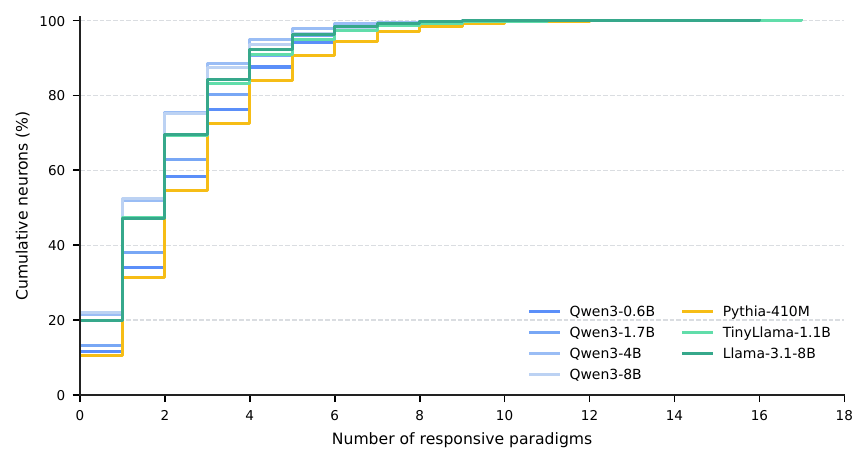}
   \caption{Cumulative distribution of paradigm-level tuning breadth across models. Even under the lenient NSI${>}0$ definition, the high-breadth tail is small: the 95th percentile is 5--7 paradigms and fewer than 1\% of neurons respond to 10 or more paradigms in every model.}
   \label{fig:model_breadth_distribution}
\end{figure}

Table~\ref{tab:model_coverage} reports architecture, calibration, sparsity, and tuning-breadth statistics across the seven checkpoints. Figure~\ref{fig:model_breadth_distribution} shows the corresponding breadth distributions.

\begin{table}[H]
   \centering
   \scriptsize
   \setlength{\tabcolsep}{3.5pt}
   \caption{Cross-model selectivity and tuning breadth over 68 paradigms. ``Paradigms $>2$'' is the number of paradigms containing any NSI${>}2$ unit. Breadth $B$ counts the number of paradigms with NSI${>}0$ for each neuron.}
   \label{tab:model_coverage}
   \begin{tabular}{lrrrrrrr}
      \toprule
      Model & Layers $\times$ width & NSI${>}0$ (\%) & NSI${>}2$ (\%) & Paradigms $>2$ & Mean $B$ & P95/max $B$ & $B{\geq}10$ (\%) \\
      \midrule
      Qwen3-0.6B     & $28\times1024$ & 3.568 & $1.54{\times}10^{-4}$ & 3 & 2.43 & 6/13 & 0.18 \\
      Qwen3-1.7B     & $28\times2048$ & 3.261 & $2.82{\times}10^{-4}$ & 6 & 2.22 & 5/13 & 0.09 \\
      Qwen3-4B       & $36\times2560$ & 2.524 & $1.12{\times}10^{-4}$ & 2 & 1.72 & 5/13 & 0.06 \\
      Qwen3-8B       & $36\times4096$ & 2.593 & $1.92{\times}10^{-3}$ & 4 & 1.76 & 5/15 & 0.24 \\
      Pythia-410M    & $24\times1024$ & 3.942 & $5.39{\times}10^{-4}$ & 2 & 2.68 & 7/16 & 0.86 \\
      TinyLlama-1.1B & $22\times2048$ & 2.939 & $6.20{\times}10^{-4}$ & 2 & 2.00 & 5/17 & 0.35 \\
      Llama-3.1-8B   & $32\times4096$ & 2.846 & $8.98{\times}10^{-5}$ & 4 & 1.94 & 5/16 & 0.11 \\
      \bottomrule
   \end{tabular}
\end{table}

\FloatBarrier
\subsection{Selectivity distributions and high-selectivity cases}
\begin{figure*}[t]
   \centering
   \includegraphics[width=1.0\linewidth]{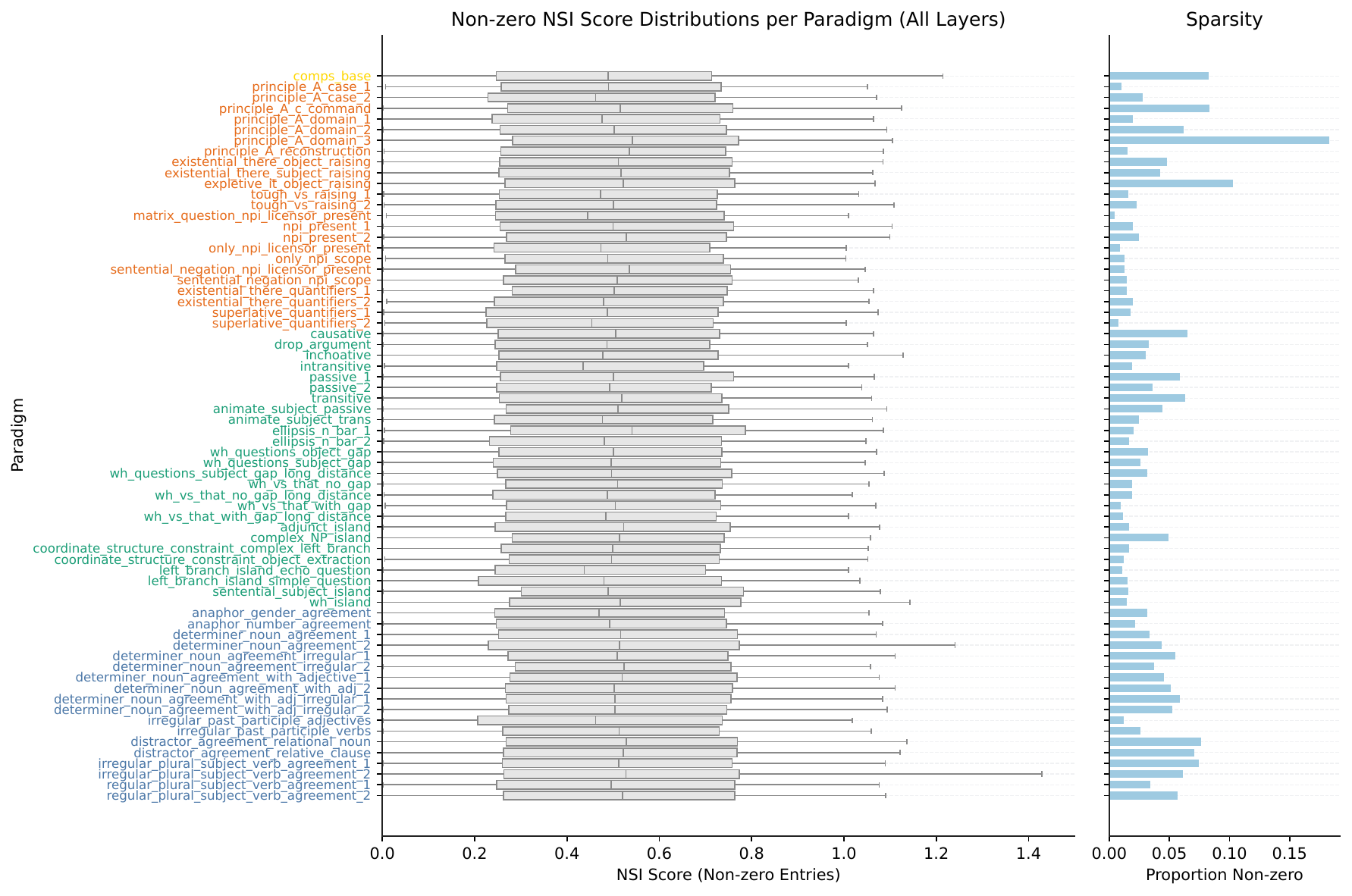}
   \caption{\textbf{Left}: Distributions of non-zero NSI scores (i.e., over sensitive neurons with $\mathrm{NSI}{>}0$) aggregated across all layers, shown separately for each paradigm. Across paradigms, the mean NSI is typically around $\sim 0.5$, indicating that most sensitive neurons exhibit only weak separability. \textbf{Right}: Paradigm-wise proportion of sensitive neurons across all 68 BLiMP/COMPS paradigms for Qwen3-0.6B.}
   \label{fig:proportion_positive}
\end{figure*}

\begin{figure*}[t]
   \centering
   \includegraphics[width=1.0\linewidth]{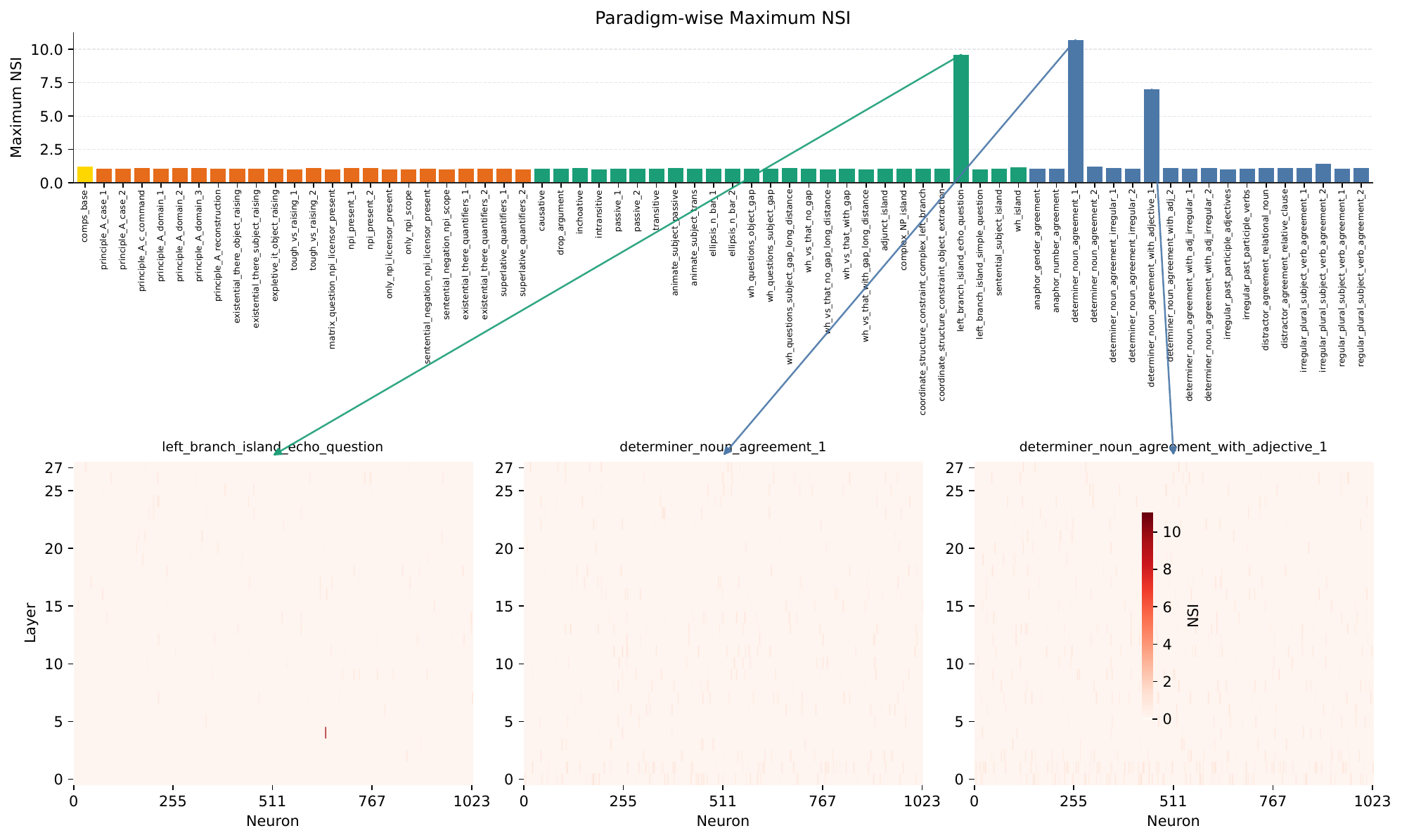}
   \caption{Paradigm-wise maximum NSI scores across all 68 BLiMP/COMPS paradigms for Qwen3-0.6B. Only three paradigms contain a unit with NSI${>}2$; the bottom panels show that each case is realized by one unit across the network.}
   \label{fig:PNS_all}
\end{figure*}

The three paradigms containing an NSI${>}2$ unit in the main 500-permutation Qwen3-0.6B analysis are illustrated below. As Figure~\ref{fig:PNS_all} shows, each paradigm contains only one such unit across all $28\times1024$ layer-neuron coordinates.

(i) \textbf{Domain:} Morphology; \textbf{Phenomenon:} Determiner--Noun Agreement\\
\hspace*{2em}\textbf{Paradigm:} Determiner--Noun Agreement 1\\
\hspace*{2em}\textit{a) \hspace*{0.5em}Raymond is selling this \underline{sketch.}} \\
\hspace*{2em}\textit{b) *Raymond is selling this \underline{sketches.}} \\

(ii) \textbf{Domain:} Syntax; \textbf{Phenomenon:} Island Effects\\
\hspace*{2em}\textbf{Paradigm:} Left-Branch Island Echo Question\\
\hspace*{2em}\textit{a) \hspace*{0.5em}Benjamin was researching whose books?} \\
\hspace*{2em}\textit{b) *Whose was Benjamin researching books?} \\

(iii) \textbf{Domain:} Morphology; \textbf{Phenomenon:} Determiner--Noun Agreement\\
\hspace*{2em}\textbf{Paradigm:} Determiner--Noun Agreement with Adjective 1\\
\hspace*{2em}\textit{a) \hspace*{0.5em}Rebecca was criticizing those good \underline{documentaries.}} \\
\hspace*{2em}\textit{b) *Rebecca was criticizing those good \underline{documentary.}} \\

\FloatBarrier
\subsection{Tuning breadth across domains and phenomena}
\begin{figure}[H]
   \centering
   \begin{minipage}{0.48\textwidth}
      \centering
      \includegraphics[width=\linewidth]{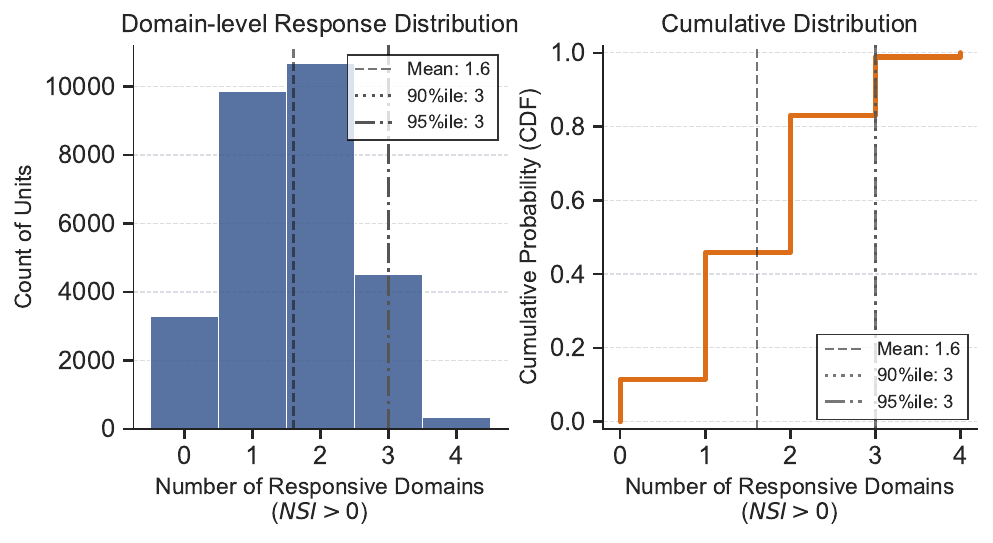}
      \caption{Domain-level tuning breadth. Most neurons respond to only one or two of the four domains.}
      \label{fig:distribution_field}
   \end{minipage}
   \hfill
   \begin{minipage}{0.48\textwidth}
      \centering
      \includegraphics[width=\linewidth]{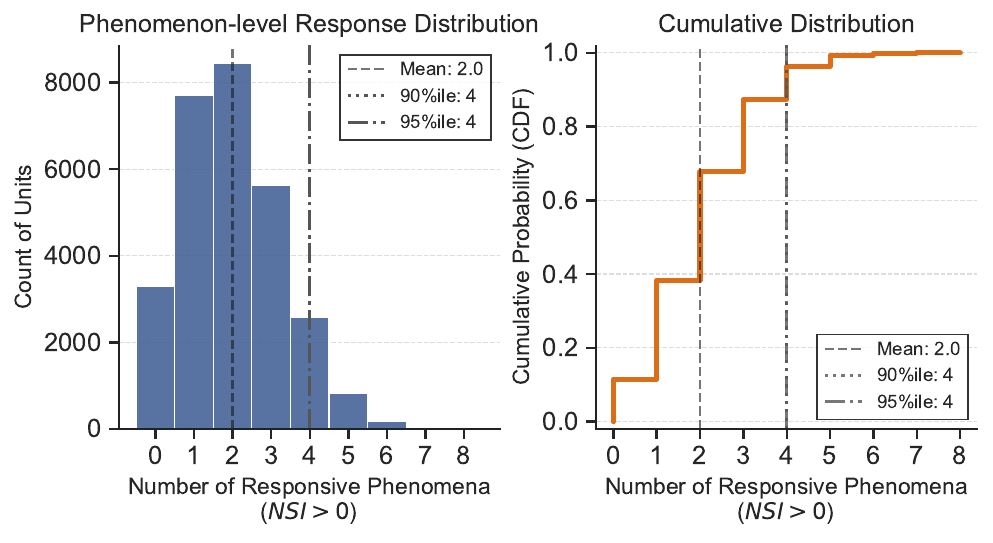}
      \caption{Phenomenon-level tuning breadth. More than 90\% of neurons respond to four or fewer phenomena.}
      \label{fig:distribution_form}
   \end{minipage}
\end{figure}

\clearpage
\FloatBarrier
\section{Similarity Metrics and Threshold Calibration}
\label{sec:threshold_calibration}

\paragraph{Threshold sweep and empirical null.}
Table~\ref{tab:threshold_sweep} reports the complete threshold sweep for Qwen3-0.6B. Weak positive effects occur in every paradigm, but the above-threshold population contracts rapidly. At NSI${>}1$, an average paradigm contains only 27.1 units out of 28,672. At every tested threshold from 1.645 to 3, only three paradigms contain any above-threshold unit, with one unit in each paradigm.

\begin{table}[H]
   \centering
   \scriptsize
   \setlength{\tabcolsep}{4pt}
   \caption{Threshold sweep over 68 paradigms and $28\times1024$ units per paradigm.}
   \label{tab:threshold_sweep}
   \begin{tabular}{rrrrr}
      \toprule
      Threshold & Units above (\%) & Paradigms with any & Mean count & Max count \\
      \midrule
      0     & 3.7010 & 68 & 1061.15 & 5229 \\
      1     & 0.0944 & 68 & 27.06   & 120 \\
      1.645 & $1.54{\times}10^{-4}$ & 3 & 0.044 & 1 \\
      2     & $1.54{\times}10^{-4}$ & 3 & 0.044 & 1 \\
      2.326 & $1.54{\times}10^{-4}$ & 3 & 0.044 & 1 \\
      2.58  & $1.54{\times}10^{-4}$ & 3 & 0.044 & 1 \\
      3     & $1.54{\times}10^{-4}$ & 3 & 0.044 & 1 \\
      \bottomrule
   \end{tabular}
\end{table}

For empirical calibration, we compute $p=(1+\#\{S^{\mathrm{perm}}\geq S^{\mathrm{raw}}\})/(1+N_{\mathrm{perm}})$ and apply Benjamini--Hochberg correction within each paradigm. No unit survives at $q<0.05$ or $q<0.01$. This result is deliberately treated as a conservative sanity check: with $N_{\mathrm{perm}}=200$, the minimum attainable empirical p-value is $1/201=0.004975$, which is too coarse for a powerful correction over 28,672 units. The permutation nulls are also non-Gaussian (mean absolute skewness $1.99$; mean paradigm-level median excess kurtosis $4.45$). These diagnostics motivate describing NSI${>}2$ as an operational strong-selectivity threshold rather than a literal Gaussian significance cutoff.

\paragraph{Alternative similarity metrics.}
We recompute the score with Spearman rank correlation and cosine similarity while keeping the 68 paradigms, Qwen3-0.6B checkpoint, layer--neuron grid, and permutation-normalization procedure fixed. At the operational strong-selectivity threshold $z>2$, only $0.222\%$ of all tested layer--neuron units exceed the threshold under Spearman and $0.277\%$ under cosine. Thus, even these more permissive variants leave more than $99.7\%$ of units below threshold, preserving the central conclusion that strong single-unit selectivity is sparse.

Neither alternative is as well matched to our estimand. Each vector coordinate in our analysis is a repeated observation of the same scalar neuron across matched items, rather than a distinct representation feature. Spearman replaces activation values with ranks, making it invariant to monotonic rescaling but discarding graded response magnitudes and becoming sensitive to ties. Cosine does not center each neuron's responses across items, so baseline and mean-level activation effects can appear as separability; its standardized scores can also become unstable when the permutation variance is extremely small. Pearson correlation instead centers each neuron's item-wise response vector and directly measures whether the relative response pattern is preserved across the two members of a minimal pair. We therefore retain Pearson for the primary analysis and treat Spearman and cosine as complementary metric checks.

\clearpage
\FloatBarrier
\section{Pairing and Deletion Controls}
\label{sec:pairing_deletion}
All controls in this section are run on Qwen3-0.6B with 200 permutations per paradigm.

\subsection{Random-pair controls}
We compare the original item-matched grammatical--ungrammatical pairs with three deranged pairings, so no sentence remains paired with itself. Table~\ref{tab:pairing_controls} separates raw decorrelation from null-normalized selectivity. Random same-label pairs have raw scores near 0.5 because unrelated sentences are weakly correlated, but they produce no NSI${>}2$ units. Conversely, cross-item good--bad pairs generate many sporadic outliers, showing that item alignment is necessary for interpreting the contrast. The original row here comes from the 200-permutation control run; its four above-threshold paradigms should not be conflated with the three paradigms in the 500-permutation main analysis.

\begin{table}[H]
   \centering
   \scriptsize
   \setlength{\tabcolsep}{3.5pt}
   \caption{Pairing controls over 68 paradigms.}
   \label{tab:pairing_controls}
   \begin{tabular}{lrrrrr}
      \toprule
      Condition & Mean raw & NSI${>}0$ (\%) & NSI${>}2$ (\%) & Paradigms $>2$ & Median max \\
      \midrule
      Matched good--bad    & 0.1438 & 3.469 & $6.67{\times}10^{-4}$ & 4  & 1.128 \\
      Random good--good    & 0.5008 & 100.000 & 0 & 0 & 0.980 \\
      Random bad--bad      & 0.5009 & 100.000 & 0 & 0 & 0.989 \\
      Cross-item good--bad & 0.5004 & 16.464 & 0.0454 & 50 & 3.325 \\
      \bottomrule
   \end{tabular}
\end{table}

\subsection{Critical-token deletion}
For agreement paradigms, we compare the original pair with deletion of either the annotated critical token or a matched non-critical token. The clean test is the 10-paradigm one-prefix subset: the critical word directly realizes the good--bad contrast, so deleting it makes the pair identical or nearly identical. Table~\ref{tab:deletion_controls} shows that critical deletion nearly eliminates raw separability, whereas random deletion does not.

\begin{table}[H]
   \centering
   \scriptsize
   \setlength{\tabcolsep}{5pt}
   \caption{Critical-token deletion controls. We emphasize raw separability because deletion changes the geometry and variance of the permutation null.}
   \label{tab:deletion_controls}
   \begin{tabular}{llrrr}
      \toprule
      Subset & Condition & Paradigms & Mean raw & Median raw \\
      \midrule
      One-prefix & Original          & 10 & 0.0753 & 0.0757 \\
                 & Critical deleted  & 10 & 0.0011 & 0.0007 \\
                 & Random deleted    & 10 & 0.0746 & 0.0772 \\
      \midrule
      Two-prefix & Original          & 6 & 0.1007 & 0.1069 \\
                 & Shared word deleted & 6 & 0.1802 & 0.1500 \\
                 & Random deleted    & 6 & 0.0971 & 0.0933 \\
      \bottomrule
   \end{tabular}
\end{table}

For example, deleting the critical verb from \textit{Paula references Robert} versus \textit{Paula reference Robert} leaves the same fragment, \textit{Paula Robert}, on both sides. In two-prefix paradigms, however, the annotated shared target is not the contrasting word: deleting it can leave prefixes such as \textit{The students} versus \textit{The student}. Those six paradigms therefore do not constitute a clean contrast-removal test and are reported separately rather than used in the main conclusion. Normalized NSI after critical deletion is not emphasized because near-identical pairs also collapse the null variance, making the resulting z-score unstable.

\clearpage
\FloatBarrier
\section{Probe, NSI, and Behavior}
\label{sec:probe_behavior}
The behavioral analysis covers the 67 standard BLiMP paradigms and excludes COMPS. For each minimal pair, behavior is correct when the grammatical sentence receives higher mean token log-probability. Whole-vector results use a logistic-regression probe on layer-14 final-token representations with five-fold stratified cross-validation. Mean behavioral accuracy is $0.758$, whereas mean probe accuracy is $0.917$.

\begin{table}[H]
   \centering
   \scriptsize
   \setlength{\tabcolsep}{4pt}
   \caption{Correlations across 67 BLiMP paradigms.}
   \label{tab:behavior_correlations}
   \begin{tabular}{lrrr}
      \toprule
      Comparison & $n$ & Pearson $r$ & Spearman $\rho$ \\
      \midrule
      Behavior vs. layer-14 probe & 67 & -0.032 & 0.023 \\
      Behavior vs. peak probe     & 67 & -0.030 & 0.008 \\
      Behavior vs. mean NSI       & 67 & -0.021 & -0.022 \\
      Behavior vs. maximum NSI    & 67 & 0.169  & 0.177 \\
      Layer-14 probe vs. mean NSI & 67 & -0.557 & -0.578 \\
      Layer-14 probe vs. maximum NSI & 67 & 0.092 & -0.178 \\
      \bottomrule
   \end{tabular}
\end{table}

\begin{figure}[H]
   \centering
   \includegraphics[width=0.92\linewidth]{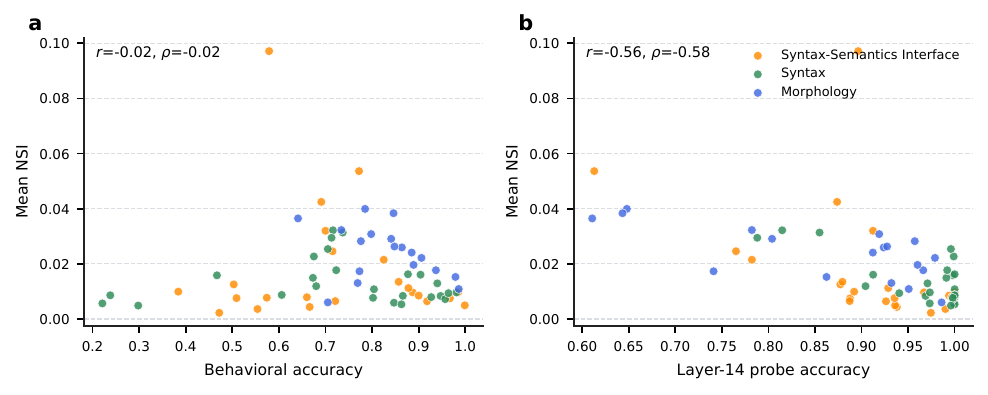}
   \caption{Additional comparisons among behavior, whole-vector probing, and neuron-level NSI. \textbf{(a)} Mean NSI is uncorrelated with behavioral accuracy. \textbf{(b)} Whole-vector probe accuracy and mean NSI are negatively correlated, further showing that distributed linear separability and average single-neuron selectivity capture different properties.}
   \label{fig:behavior_nsi_supplement}
\end{figure}

\begin{table}[H]
   \centering
   \scriptsize
   \setlength{\tabcolsep}{4pt}
   \caption{Domain-level averages over the 67 BLiMP paradigms. The legacy \texttt{semantics}, \texttt{syntax\_semantics}, and \texttt{syntax/semantics} source labels are merged into the Syntax--Semantics Interface domain.}
   \label{tab:behavior_fields}
   \begin{tabular}{lrrrrr}
      \toprule
      Domain & Paradigms & Behavior & Probe & Mean NSI & Max NSI \\
      \midrule
      Syntax--Semantics Interface & 23 & 0.727 & 0.907 & 0.018 & 1.133 \\
      Syntax                      & 26 & 0.736 & 0.964 & 0.014 & 1.424 \\
      Morphology                  & 18 & 0.831 & 0.861 & 0.024 & 1.932 \\
      \bottomrule
   \end{tabular}
\end{table}

\begin{table}[H]
   \centering
   \scriptsize
   \setlength{\tabcolsep}{5pt}
   \caption{Representative paradigms ranked by behavioral accuracy. Near-perfect linear probing can coexist with poor model behavior.}
   \label{tab:behavior_tasks}
   \begin{tabular}{llrrrr}
      \toprule
      Paradigm & Domain & Behavior & Probe & Mean NSI & Max NSI \\
      \midrule
      \texttt{principle\_A\_case\_1} & Syntax--Semantics Interface & 0.999 & 0.936 & 0.005 & 1.134 \\
      \texttt{anaphor\_number\_agreement} & Morphology & 0.986 & 0.951 & 0.011 & 1.118 \\
      \texttt{principle\_A\_domain\_1} & Syntax--Semantics Interface & 0.985 & 1.000 & 0.010 & 1.224 \\
      \midrule
      \texttt{wh\_vs\_that\_with\_gap\_long\_distance} & Syntax & 0.221 & 0.973 & 0.006 & 1.098 \\
      \texttt{sentential\_subject\_island} & Syntax & 0.238 & 1.000 & 0.009 & 1.105 \\
      \texttt{wh\_vs\_that\_with\_gap} & Syntax & 0.298 & 0.996 & 0.005 & 1.064 \\
      \texttt{npi\_present\_1} & Syntax--Semantics Interface & 0.384 & 0.892 & 0.010 & 1.165 \\
      \texttt{drop\_argument} & Syntax & 0.467 & 0.998 & 0.016 & 1.127 \\
      \bottomrule
   \end{tabular}
\end{table}

\clearpage
\FloatBarrier
\section{Targeted Ablation}
\label{sec:targeted_ablation}

We use ablation as a targeted causal follow-up rather than as a second neuron-selection procedure. The observational NSI analysis is frozen before intervention and identifies one NSI${>}2$ coordinate in each of three Qwen3-0.6B paradigms: layer 20, neuron 389 for \texttt{determiner\_noun\_agreement\_1} (NSI $10.073$); layer 20, neuron 389 for \texttt{determiner\_noun\_agreement\_with\_adjective\_1} (NSI $6.037$); and layer 4, neuron 646 for \texttt{left\_branch\_island\_echo\_question} (NSI $8.927$). Selection and evaluation use the same 1,000 minimal pairs, so this is a within-benchmark intervention test, not an independent replication.

At the frozen target layer, we zero the top $k\in\{1,5,10,20\}$ residual-stream dimensions at every non-padding token; each top group contains the above-threshold candidate together with its highest-scoring same-layer neighbors. The main analysis treats $k\in\{5,10,20\}$ as group interventions; $k=1$ is an auxiliary necessity check. Signed null-normalized Pearson scores determine the top and bottom sets. Random controls comprise 100 unique same-layer sets per $k$, sampled from a pool that excludes the union of the top-20 and bottom-20 dimensions. All full runs use bfloat16. The primary outcome is the change in the mean total log-probability margin, $M=\log p(x^+)-\log p(x^-)$. We obtain item-level 95\% CIs from 2,000 paired bootstrap resamples. Separately, the one-sided random-control statistic is
\begin{equation}
   p_{\mathrm{rand}}=\frac{1+\#\{\Delta M_{\mathrm{random}}\leq\Delta M_{\mathrm{top}}\}}{101},
\end{equation}
where a more negative $\Delta M$ is considered more damaging. Bootstrap resampling is used only for confidence intervals, not for this empirical $p$-value.

\begin{figure*}[t]
   \centering
   \includegraphics[width=\textwidth]{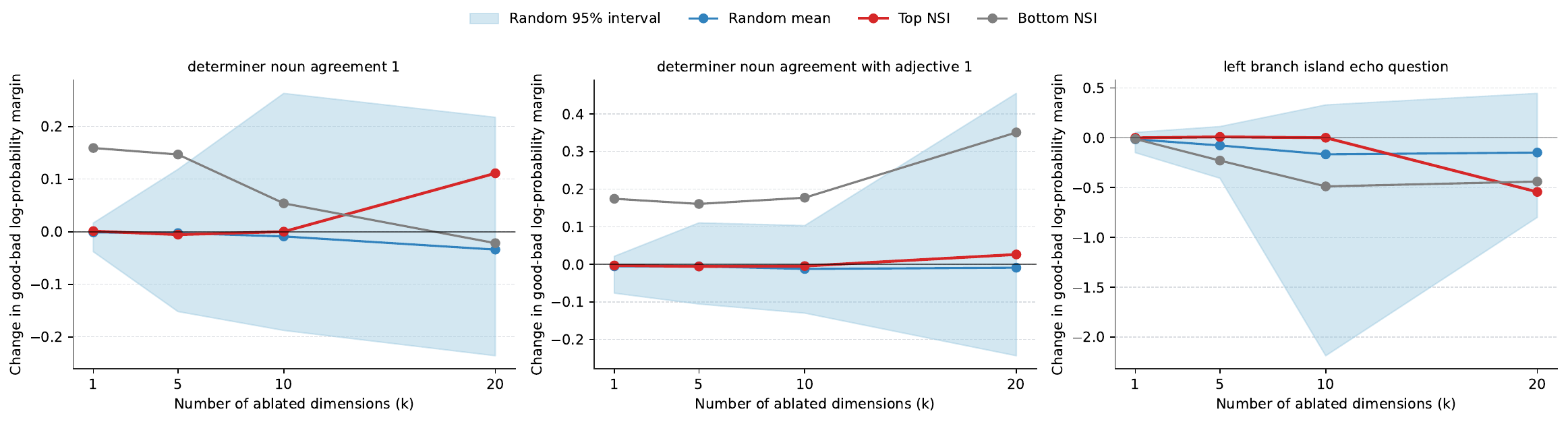}
   \caption{Targeted ablation at each frozen candidate's actual layer. Red and gray lines show the top-score and bottom signed-score sets; the blue line and band show the mean and 2.5th--97.5th percentile interval across 100 same-layer random sets. Negative values indicate a reduced grammatical-over-ungrammatical log-probability margin.}
   \label{fig:targeted_ablation}
\end{figure*}

\begin{table*}[t]
   \centering
   \scriptsize
   \setlength{\tabcolsep}{3.5pt}
   \caption{Complete targeted-ablation results. Top CI is the paired-bootstrap 95\% interval over items; Random interval is the 2.5th--97.5th percentile interval across 100 random sets. $\Delta A$ is the change in total-log-probability minimal-pair accuracy.}
   \label{tab:ablation_full}
   \resizebox{\textwidth}{!}{%
   \begin{tabular}{@{}lrrrrrrrr@{}}
      \toprule
      Paradigm & $k$ & Top $\Delta M$ & Top CI & Random mean & Random interval & Bottom $\Delta M$ & $\Delta A$ & $p_{\mathrm{rand}}$ \\
      \midrule
      Det.--noun agr. & 1  &  0.001 & [$-0.004, 0.006$] & $-0.001$ & [$-0.038, 0.017$] &  0.159 &  0.001 & 0.554 \\
                       & 5  & $-0.005$ & [$-0.014, 0.003$] & $-0.003$ & [$-0.152, 0.119$] &  0.147 & $-0.001$ & 0.446 \\
                       & 10 &  0.000 & [$-0.011, 0.011$] & $-0.009$ & [$-0.188, 0.263$] &  0.054 &  0.002 & 0.634 \\
                       & 20 &  0.111 & [$ 0.084, 0.138$] & $-0.034$ & [$-0.236, 0.218$] & $-0.022$ & $-0.002$ & 0.911 \\
      \midrule
      Det.--noun agr.+adj. & 1  & $-0.003$ & [$-0.008, 0.001$] & $-0.005$ & [$-0.077, 0.021$] & 0.174 &  0.001 & 0.396 \\
                            & 5  & $-0.006$ & [$-0.015, 0.004$] & $-0.005$ & [$-0.106, 0.110$] & 0.161 &  0.000 & 0.545 \\
                            & 10 & $-0.005$ & [$-0.020, 0.012$] & $-0.012$ & [$-0.130, 0.103$] & 0.177 & $-0.003$ & 0.604 \\
                            & 20 &  0.026 & [$ 0.004, 0.048$] & $-0.009$ & [$-0.243, 0.455$] & 0.351 & $-0.003$ & 0.733 \\
      \midrule
      Left-branch island & 1  & $-0.0003$ & [$-0.018, 0.019$] & $-0.015$ & [$-0.149, 0.054$] & $-0.012$ & $-0.003$ & 0.614 \\
                          & 5  &  0.009 & [$-0.019, 0.038$] & $-0.077$ & [$-0.406, 0.113$] & $-0.229$ & $-0.004$ & 0.752 \\
                          & 10 &  0.001 & [$-0.039, 0.040$] & $-0.166$ & [$-2.187, 0.330$] & $-0.488$ & $-0.006$ & 0.693 \\
                          & 20 & $-0.543$ & [$-0.630,-0.455$] & $-0.148$ & [$-0.799, 0.446$] & $-0.439$ & $-0.009$ & 0.109 \\
      \bottomrule
   \end{tabular}
   }
\end{table*}

Figure~\ref{fig:targeted_ablation} and Table~\ref{tab:ablation_full} summarize the intervention results. Across the nine primary group comparisons ($k\in\{5,10,20\}$), no targeted set is more damaging than same-size random controls at $p<0.05$. Although the left-branch top-20 intervention has a clear negative item-level effect, it is not extreme relative to the random interventions, and its bottom control is similarly negative. The auxiliary top-1 checks are also small and non-selective, but we do not treat single-coordinate ablation as the main causal test. We therefore conclude only that this intervention does not identify a behaviorally privileged high-NSI group, not that the selected dimensions are causally irrelevant. Redundancy, superposition, and the coarseness of residual-coordinate zeroing remain possible explanations for the lack of selective effects.
\FloatBarrier

\end{document}